\documentclass[lettersize,journal]{IEEEtran}
\usepackage{amsmath,amsfonts}
\usepackage{algorithmic}
\usepackage{algorithm}
\usepackage{array}
\usepackage[caption=false,font=normalsize,labelfont=sf,textfont=sf]{subfig}
\usepackage{textcomp}
\usepackage{stfloats}
\usepackage{url}
\usepackage{verbatim}
\usepackage{graphicx}
\usepackage{cite}
\usepackage{booktabs}
\usepackage{makecell}
\usepackage{multirow}
\usepackage{soul, color, xcolor}
\usepackage{amsmath,amssymb,amsfonts}
\usepackage{xcolor}
\usepackage{colortbl}
\usepackage{balance}
\usepackage[T1]{fontenc}
\usepackage[utf8]{inputenc}
\usepackage{newtxtext,newtxmath}

\usepackage{xcolor}
\usepackage{listings}
\usepackage[most]{tcolorbox}

\definecolor{JSONBlue}{HTML}{2563EB}
\definecolor{JSONBack}{HTML}{F8FAFC}
\definecolor{JSONFrame}{HTML}{CBD5E1}
\definecolor{JSONString}{HTML}{047857}
\definecolor{JSONKey}{HTML}{1E3A8A}
\definecolor{JSONKeyword}{HTML}{B45309}

\lstdefinelanguage{json}{
    basicstyle=\ttfamily\tiny\linespread{0.82}\selectfont,
    showstringspaces=false,
    breaklines=true,
    breakatwhitespace=false,
    columns=fullflexible,
    keepspaces=true,
    morestring=[b]",
    stringstyle=\color{JSONKey},
    morekeywords={true,false,null},
    keywordstyle=\color{JSONKeyword}\bfseries
}

\newtcblisting{compactjson}[2][]{
    enhanced,
    breakable,
    listing only,
    listing engine=listings,
    colback=JSONBack,
    colframe=JSONFrame,
    boxrule=0.45pt,
    arc=1mm,
    boxsep=1pt,
    left=2pt,
    right=2pt,
    top=2pt,
    bottom=2pt,
    title={#2},
    fonttitle=\bfseries\scriptsize,
    coltitle=white,
    attach boxed title to top left={xshift=1mm,yshift=-1.5mm},
    boxed title style={
        colback=JSONBlue,
        colframe=JSONBlue,
        boxrule=0pt,
        arc=0.8mm,
        left=1.5mm,
        right=1.5mm,
        top=0.4mm,
        bottom=0.4mm
    },
    listing options={
        language=json,
        basicstyle=\ttfamily\scriptsize\linespread{1}\selectfont,
        numbers=none,
        breaklines=true,
        columns=fullflexible,
        keepspaces=true,
        showstringspaces=false
    },
    #1
}

\begin{document}

\title{AutoMCU: Feasibility-First MCU Neural Network Customization via LLM-based Multi-Agent Systems}

\author{Penglin~Dai,~\IEEEmembership{Member,~IEEE}, Zijie~Zhou, Xincao~Xu,~\IEEEmembership{Member,~IEEE,}
Junhua~Wang,~\IEEEmembership{Member,~IEEE,} Xiao~Wu,~\IEEEmembership{Member,~IEEE,} Lixin Duan,~\IEEEmembership{Member,~IEEE}
\thanks{This work was supported in part by the National Natural Science Foundation of China under Grant 62172342; in part by the Guangdong Basic and Applied Basic Research Foundation under Grant 2025A1515012825; in part by Yibin Science and Technology Program under Grant 2025JC014, and in part by Science, Technology and Innovation Project of Shenzhen Longhua District under Grant 20260309G23410662.}
\thanks{P. Dai, Z. Zhou, X. Wu are with the School of Computing and Artificial Intelligence, Southwest Jiaotong University, Chengdu 611756, China. (email: penglindai@swjtu.edu.cn; zzj1921@my.swjtu.edu.cn; wuxiaohk@swjtu.edu.cn).}%
\thanks{X. Xu, L, Duan are with the Shenzhen Institute for Advanced Study, University of Electronic Science and Technology of China, Shenzhen 518110, China. (email: xc.xu@uestc.edu.cn; lxduan@uestc.edu.cn.)}
\thanks{J. Wang is with the School of Computer Science and Engineering, Northeastern University, Shenyang 110819, China. (email: wangjunhua@cse.neu.edu.cn)}}

\markboth{Journal of \LaTeX\ Class Files,~Vol.~14, No.~8, August~2021}%
{Shell \MakeLowercase{\textit{et al.}}: A Sample Article Using IEEEtran.cls for IEEE Journals}


\maketitle

\begin{abstract}
Deploying neural networks on microcontroller units (MCUs) is critical for edge intelligence but remains challenging due to tight memory, storage, and computation constraints. Existing approaches, such as model compression and hardware-aware neural architecture search (HW-NAS), often depend on proxy metrics, incur high search cost, and do not fully bridge the gap between architecture design and verified deployment. This paper presents AutoMCU, a feasibility-first large language model (LLM)-based multi-agent system for automated neural network customization under MCU constraints. Given natural-language task requirements and hardware specifications, AutoMCU iteratively generates structured architecture candidates, filters infeasible designs through vendor toolchain feedback before training, evaluates feasible models under a controlled protocol, and verifies deployability through backend-grounded deployment analysis. AutoMCU includes two key mechanisms: 1) hardware-in-the-loop architecture generation for early elimination of undeployable candidates under RAM and Flash constraints, and 2) state-isolated multi-agent scheduling for stable coordination of proposal, training, evaluation, and deployment stages. Experiments on CIFAR-10 and CIFAR-100 under strict MCU constraints show that AutoMCU achieves competitive accuracy while reducing customization time to about 1--2 hours, compared with hundreds of GPU hours for representative MCU-oriented HW-NAS baselines. Comparisons with ColabNAS and the LLM-based NAS method GENIUS on NAS-Bench-201 further demonstrate the effectiveness and stability of AutoMCU. Real-device deployments on multiple STM32 microcontrollers validate its practical applicability to MCU-scale edge intelligence.
\end{abstract}


\section{Introduction}
\IEEEPARstart{W}{ith} the rapid proliferation of Internet of Things (IoT) devices, deploying Neural Networks (NNs) directly on Microcontroller Units (MCUs) has emerged as a critical enabler for ubiquitous edge intelligence, supporting a wide range of applications such as computer vision \cite{garcia2021lspnet, he2016deep}, speech recognition \cite{chan2016listen, chorowski2015attention}, and natural language processing \cite{bahdanau2014neural}. This paradigm is particularly attractive for IoT scenarios, including smart homes, industrial automation, healthcare, and agriculture, where low latency, privacy preservation, and energy efficiency are critical. However, MCUs are severely constrained in terms of computational capability, on-chip memory, and non-volatile storage, making the deployment of modern NN models highly challenging.

\begin{figure}[t]
  \centering
  \includegraphics[width=0.485\textwidth]{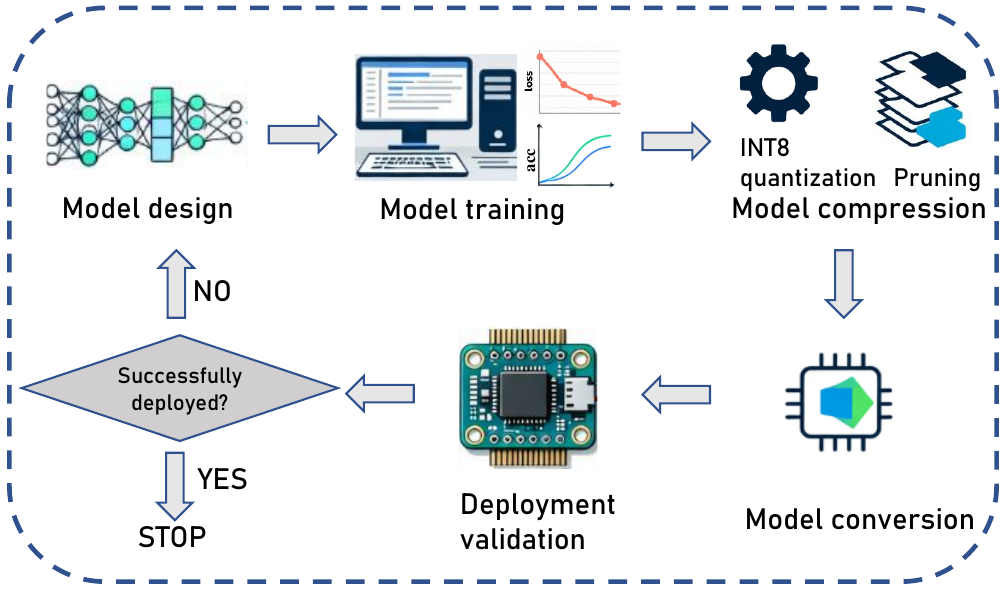}
  \caption{The general process for neural network customization for MCU.}
  \label{Fig: mcu deploy}
\end{figure}

To satisfy these constraints, NN models must be carefully customized to match the resource budget of a target MCU. Existing techniques such as pruning, quantization, and knowledge distillation can reduce model size and computational cost, but deploying an NN on a concrete MCU platform still typically requires repeated manual iteration across architecture design, training, model conversion, and hardware validation, as shown in Fig. \ref{Fig: mcu deploy}. Developers must reason about RAM and Flash limits, choose architecture configurations, train candidate models, and test whether they can be converted and executed by vendor-specific backends such as TFLite Micro \cite{david2021tensorflow} or STM32Cube.AI. In practice, models that appear efficient under proxy metrics such as parameter count or MACs may still fail deployment because of unsupported operators, backend-specific memory allocation behavior, or conversion/runtime constraints. This trial-and-error workflow significantly increases development cost and raises the barrier to deploying NNs on MCUs.
A large body of prior work has explored automated or semi-automated model design for resource-constrained devices. Compression and lightweight design methods reduce model complexity through pruning, quantization, distillation, or manually crafted efficient architectures \cite{han2015deep,jacob2018quantization,wang2019haq,deng2020model}.
Neural Architecture Search (NAS) and Hardware-aware NAS (HW-NAS) further automate architecture optimization under accuracy and efficiency objectives\cite{zoph2017neural,real2017large,pham2018efficient,guo2020single,he2021automl,lin2020mcunet,tan2019mnasnet,chu2021fairnas,wu2019fbnet,liberis2021munas}.More recently, Large Language Models (LLMs) have also been explored for architecture generation and search-space exploration\cite{zheng2023can}. While these directions have substantially advanced automated model design, important challenges remain for deployment-oriented MCU customization. In particular, many existing approaches optimize proxy hardware metrics or predefined search objectives rather than directly prioritizing backend-verified deployability under hard memory and operator constraints. Moreover, architecture search, training, and deployment validation are often handled as separate stages, leaving substantial manual effort in the overall workflow.

To address this gap, we propose AutoMCU, a feasibility-first LLM-based multi-agent system (MAS) for end-to-end neural network customization under microcontroller constraints. AutoMCU targets a complementary and practically important setting for rapid discovery of deployable models under hard MCU constraints, where backend-specific feasibility is difficult to characterize analytically. Rather than performing unconstrained architecture generation or globally optimizing the full accuracy and efficiency Pareto frontier, AutoMCU couples constrained candidate proposal, controlled training, and backend-verified deployment validation in a closed loop. The LLM is used to generate and refine structured candidates conditioned on explicit MCU constraints and summarized historical feedback, while modular orchestration coordinates proposal, training, evaluation, and conversion stages through stable structured interfaces.

The main contributions of this work are summarized as follows:
\begin{itemize}
\item We shift MCU neural network customization from proxy-driven hardware-aware search to deployment-grounded feasibility-first search, where backend-verified deployability is used as an iterative feedback signal rather than only a final validation step.
\item We develop a hardware-in-the-loop architecture generation mechanism that constrains LLM-generated candidates into structured and constructible architecture specifications, and filters infeasible designs through vendor backend analysis before training to avoid wasted optimization on undeployable models.
\item We design a supervisor-driven multi-agent orchestration mechanism with state-isolated agents and structured summary exchange, enabling stable long-horizon automation across proposal, training, evaluation, conversion, and history-guided refinement stages.
\item We conduct empirical evaluations against MCU-oriented NAS baselines and recent LLM-based architecture search methods, including $\mu$NAS, \cite{liberis2021munas}, ColabNAS\cite{garavagno2024colabnas} and GENIUS\cite{zheng2023can}. The results, together with ablation studies, demonstrate that backend-verified feasibility checks, historical feedback, and modular orchestration improve the efficiency and stability of finding deployable MCU neural networks.
\end{itemize}

The remainder of this paper is organized as follows:  Section \ref{sec:related_work} reviews the related work. Section \ref{sec:system_model} presents the AutoMCU system. Section \ref{sec:performance_evaluation} evaluates the performance.  Finally, Section \ref{sec:Conclusion} gives a conclusion and discusses the future work.

\section{Related Work}\label{sec:related_work}
\subsection{Model Compression and Lightweight Design Methods}
Traditional approaches for deploying neural networks on resource-constrained MCU platforms often rely on manually designed lightweight architectures, such as MobileNet\cite{sandler2018mobilenetv2} and ShuffleNet\cite{ma2018shufflenet}, together with model compression techniques\cite{liang2023collaborative,liu2020adadeep}, including pruning\cite{han2015learning,luo2017thinet,he2017channel,cha2025target} and quantization\cite{han2015deep,jacob2018quantization,wang2019haq,11049020}.  Lightweight architectures reduce computational cost through operator-level design choices such as depthwise separable convolutions, channel shuffling, and grouped operations, while compression methods further decrease model size and arithmetic complexity by removing redundant parameters or lowering numerical precision.

Although these techniques can alleviate resource pressure to some extent, they are usually developed for fixed or relatively homogeneous hardware settings and lack adaptability to the highly fragmented and diverse resource constraints of MCU devices. In practice, achieving a deployable model often requires expert engineers to manually and repeatedly adjust network depth, channel width, and compression strategies for each target device. Moreover, compression is commonly applied as a post-processing step after model training, making it difficult to integrate seamlessly with hardware-aware evaluation and deployment toolchains. As a result, these methods are costly, expertise-intensive, and poorly suited for fully automated, end-to-end model customization workflows in MCU scenarios.

\subsection{NAS and HW-NAS}
NAS provides a systematic approach to automating neural network design, and hardware-aware NAS extends this paradigm by explicitly incorporating device-related metrics such as latency, memory footprint, or energy consumption into the search objective. Early hardware-aware approaches predominantly relied on evolutionary algorithms\cite{sinha2021evolving,dai2019chamnet}or reinforcement learning\cite{zoph2017neural,zoph2018learning,tan2019mnasnet} to explore the architecture space under efficiency constraints. These methods demonstrated the potential of automatically balancing accuracy and hardware cost, but they often require evaluating many candidate architectures, leading to substantial search overhead.

To improve search efficiency, weight-sharing and supernet-based methods\cite{cai2020once,liu2018darts,xu2019pc,cai2018proxylessnas} enable many candidate sub-networks to share parameters during training. This paradigm significantly reduces search cost, but it also introduces coupling between sub-networks and typically requires a carefully predefined search space, which can limit flexibility under diverse MCU constraints. Training-free or zero-shot NAS methods\cite{li2024zero,lopes2021epe,lin2021zen} further reduce cost by ranking architectures using analytically derived indicators computed without full training. However, their effectiveness depends on the correlation between such indicators and downstream performance, which may vary across datasets, architectures, and deployment settings.

More closely related to our target scenario, recent works have explored resource-constrained and deployment-aware model search for compact devices, including MCU-oriented frameworks such as $\mu$NAS\cite{liberis2021munas} and ColabNAS\cite{garavagno2024colabnas}. These methods have advanced efficient model design for constrained platforms, but many still optimize estimated hardware metrics, predefined resource objectives, or bounded search spaces rather than directly prioritizing backend-verified deployability through a concrete MCU toolchain. In addition, architecture search, model training, and deployment validation are often not fully integrated into a single closed-loop workflow. Consequently, practical deployment on a target MCU may still require additional manual iteration to resolve operator incompatibilities, memory overflows, or conversion failures.

\subsection{LLM-Assisted Architecture Design and Search}
Recent studies have begun to explore the use of LLMs for neural architecture generation, search-space exploration, and candidate refinement\cite{zheng2023can}. Compared with conventional search controllers, LLMs offer a flexible interface for generating structured design proposals from textual instructions, historical feedback, and explicit constraints. This emerging direction suggests that language-guided search can complement traditional NAS pipelines, especially in settings where design requirements are difficult to encode as fixed search rules.

However, existing LLM-based architecture design and search methods are generally not developed for MCU deployment. Most of them focus on architecture generation quality or search efficiency in conventional deep learning settings, without explicitly modeling hard Flash/SRAM budgets, operator support constraints, or backend-specific conversion behavior. More importantly, they typically do not close the loop with deployment backends to verify whether generated candidates are actually executable on a target MCU platform. This limits their applicability to deployment-oriented MCU customization, where backend-verified feasibility is often as critical as model accuracy.

\subsection{Positioning of This Work}
AutoMCU is complementary to the above lines of work. Rather than replacing compression-centric or supernet-based methods for globally optimizing the full accuracy--efficiency frontier, AutoMCU focuses on rapid neural network customization under hard MCU constraints, with the goal of discovering deployable models efficiently. Its central design choice is to combine constrained candidate proposal with backend-verified feedback in a closed loop, so that infeasible candidates can be identified through real deployment validation rather than proxy estimates alone. In this framework, the LLM is primarily used for structured proposal generation and refinement, while the modular orchestration strategy supports stable automation across proposal, training, evaluation, and deployment stages.

\section{AutoMCU System}\label{sec:system_model}
\subsection{Overview of AutoMCU}
\begin{figure*}[t]
  \centering
  \includegraphics[width=1\textwidth]{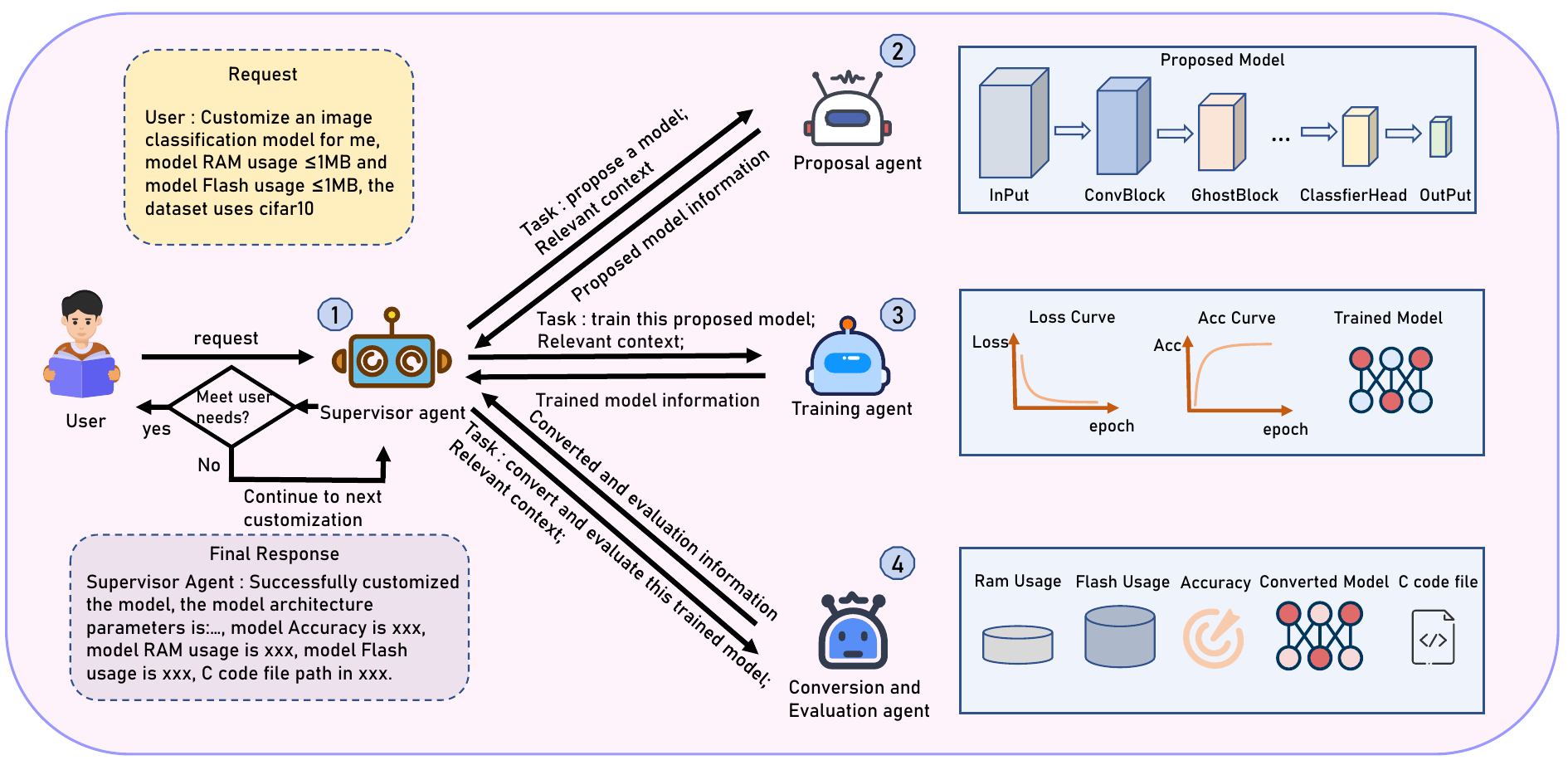}
  \caption{Overview of AutoMCU. The workflow of AutoMCU consists of four stages: 1) Task Coordination and Scheduling: the Supervisor Agent decomposes the user request into subtasks and distributes them to various expert agents. 2) Model Proposal: the Proposal Agent suggests models based on task requirements. 3) Model Training: the Training Agent trains feasible candidate models under a controlled evaluation protocol. 4) Deployment Evaluation: the Evaluation and Conversion Agent verifies hardware feasibility and prepares deployment-ready artifacts for MCUs.}
  \label{Fig:automcu}
\end{figure*}

We propose AutoMCU, a feasibility-first LLM-based multi-agent system for automated neural network customization under microcontroller constraints. Given task requirements and target hardware constraints, AutoMCU iteratively proposes candidate architectures, filters infeasible designs through a deployment backend, trains feasible candidates under a controlled protocol, and validates their deployability using vendor toolchains.
Users specify task objectives and hardware constraints (e.g., target dataset, RAM budget, and Flash budget) through natural language. AutoMCU then organizes the customization process as a closed loop consisting of candidate proposal, training, backend-based evaluation, and iterative refinement. The LLM is primarily used to generate and refine structured candidate proposals, while modular scheduling is used to coordinate the execution of different stages and maintain stable long-horizon automation.
As shown in Fig. \ref{Fig:automcu}, AutoMCU is implemented using a Supervisor component together with three task-specific modules:
\begin{itemize}
    \item \textbf{Proposal Agent}: generates and refines candidate model structures within a constrained and constructible design space;
    \item \textbf{Training Agent}: trains feasible candidate models and reports task-level performance;
    \item \textbf{Evaluation and Conversion Agent}: verifies deployment feasibility using vendor toolchains and generates deployment-ready artifacts.
\end{itemize}

The system operates in a closed-loop manner. The Proposal Agent first generates a candidate architecture, which is screened by the Evaluation and Conversion Agent against hardware constraints using vendor toolchains. Only candidates that satisfy feasibility constraints proceed to the Training Agent for performance evaluation. The resulting performance and deployment summaries are then fed back to guide subsequent proposal refinement. Two design principles underpin AutoMCU. First, hardware feasibility is treated as a hard constraint enforced before training, rather than a soft objective optimized jointly with accuracy. Second, each module maintains isolated internal state and communicates through structured summaries, which improves stability during long-horizon optimization.

\subsection{Hardware-in-the-Loop Architecture Generation}
A key challenge in MCU-oriented model customization is that architecture candidates that appear promising at a high level may still be undeployable on the target device due to strict RAM, Flash, or operator-support constraints. Traditional search methods often discover such violations only after training, which wastes substantial computation on infeasible candidates. To address this issue, AutoMCU adopts a hardware-in-the-loop architecture generation (HAG) strategy in which candidate architectures are generated in a structured form and screened by the deployment backend before training.

Rather than producing unrestricted textual model descriptions, the Proposal Agent outputs each candidate as a structured architecture specification. Formally, a candidate model is represented as
\begin{equation}
a = (\mathcal{B}, \mathcal{H}),
\label{eq:arch}
\end{equation}
where $\mathcal{B} = [l_1, l_2, \dots, l_n]$ denotes an ordered backbone layer sequence and $\mathcal{H}$ denotes the task-specific prediction head. Each layer is represented as
\begin{equation}
l_i = (t_i, \theta_i),
\label{eq:layer}
\end{equation}
where $t_i$ is the atomic module type and $\theta_i$ is the associated parameter set. The module type is restricted to a predefined set of supported building blocks, while the parameter set specifies configuration fields such as channel dimensions, kernel size, stride, padding, expansion ratio, and activation-related options.

To ensure that each candidate can be reliably instantiated and analyzed, AutoMCU implements a library of commonly used atomic neural modules (e.g., convolution blocks and depthwise convolution blocks) together with a general-purpose model builder. The builder parses the structured specification and instantiates it into an executable neural network model. A candidate is considered constructible only when its layer entries satisfy the schema constraints and inter-layer tensor dimensions remain consistent. In this way, the proposal stage is restricted to a constructible architecture space consisting of schema-valid and dimension-consistent model specifications rather than unrestricted free-form model descriptions. A simplified example of the structured architecture specification is shown below:

\begin{compactjson}{\texttt{candidate\_architecture\_example.json}}
{
  "backbone": {
    "layer_1": {"type":"conv", "in_channels":3, "out_channels":6, "kernel_size":3, "stride":1, "padding":1, "use_bn":true},
    "layer_2": {"type":"depthwise", "in_channels":6, "out_channels":6, "kernel_size":3, "stride":1, "padding":1},
    "layer_3": {"type":"downsample", "in_channels":6, "out_channels":12},
    "layer_4": {"type":"depthwise", "in_channels":12, "out_channels":12, "kernel_size":3, "stride":1, "padding":1},
    "layer_5": {"type":"downsample", "in_channels":12, "out_channels":24},
    "layer_6": {"type":"pointwise", "in_channels":24, "out_channels":24, "use_bn":true}
  },
  "head": {"type":"classifier", "num_classes":10}
}
\end{compactjson}

To improve search efficiency, AutoMCU uses a feasibility filtering step before training. Once a candidate model is instantiated, the evaluation module invokes the vendor toolchain to obtain deployment-related measurements, including RAM usage, Flash usage, and operator compatibility. Candidates that violate hard hardware constraints are rejected early and do not enter training. As a result, the search process is biased toward feasible regions of the design space, reducing wasted training on undeployable models.

To support iterative refinement, the system maintains a centralized historical evaluation repository of previously explored candidates. Each record can be written as
\begin{equation}
r_j = (\texttt{id}_j, a_j, p_j, m_j),
\label{eq:record}
\end{equation}
where $\texttt{id}_j$ denotes the candidate identifier, $a_j$ is the structured architecture specification, $p_j$ summarizes task-level performance, and $m_j$ summarizes backend-verified deployment measurements. In implementation, each repository record stores fixed fields such as the model identifier, architecture specification, and measured metrics including accuracy, RAM usage, and Flash usage.For compactness in the implementation example, task-level performance and deployment-related measurements are grouped under a unified \texttt{metrics} field. A simplified example of the historical repository format is shown below:

\begin{compactjson}{\texttt{historical\_repository\_example.json}}
[
  {
    "model_id": "model_001",
    "architecture_spec": {
      "backbone": {
        "layer_1": {"type":"conv", "in_channels":3, "out_channels":8, "kernel_size":3, "stride":1, "padding":1, "use_bn":true},
        "layer_2": {"type":"depthwise", "in_channels":8, "out_channels":8, "kernel_size":3, "stride":1, "padding":1},
        "layer_3": {"type":"downsample", "in_channels":8, "out_channels":16},
        "layer_4": {"type":"ghost", "in_channels":16, "out_channels":16, "kernel_size":3, "ratio":2, "dw_size":3},
        "layer_5": {"type":"bottleneck", "in_channels":16, "out_channels":16, "expansion":1},
        "layer_6": {"type":"downsample", "in_channels":16, "out_channels":32},
        "layer_7": {"type":"pointwise", "in_channels":32, "out_channels":32, "use_bn":true},
        "layer_8": {"type":"ghost", "in_channels":32, "out_channels":32, "kernel_size":3, "ratio":2, "dw_size":3}
      },
      "head": {"type":"classifier", "num_classes":10}
    },
    "metrics": {"model_acc":77.53, "model_ram_KB":44.11, "model_flash_KB":64.41}
  },
  ...
]
\end{compactjson}

By comparing architectural structures with their observed performance and resource usage, the Proposal Agent can refine subsequent candidates toward more favorable accuracy-resource trade-offs within the feasible deployment region.

It is worth noting that the Proposal Agent is a modular component rather than a hard-coded requirement of the framework. It can be replaced by alternative proposal strategies such as random search or heuristic rules, which also supports the ablation setting used later in the experiments. Compared with traditional HW-NAS methods that operate in fixed discrete search spaces, AutoMCU instead constrains candidate generation through structured constructibility and backend-based feasibility checks, while still allowing flexible adjustment of depth, width, and operator composition.

\subsection{Discriminative Evaluation via Controlled Training}
The goal of training within AutoMCU is not to fully optimize every candidate to its absolute best performance, but to provide discriminative and comparable feedback for architecture selection. Fully training every explored candidate to convergence would be prohibitively expensive during iterative customization. Therefore, AutoMCU adopts a controlled training strategy in which feasible candidates are evaluated under a unified training protocol.

For each candidate that passes feasibility screening, the model builder instantiates the corresponding network from its structured specification, and the Training Agent executes model training under standardized settings, including the same dataset split, preprocessing pipeline, optimizer configuration, and evaluation metric. Each candidate is trained for a limited budget to obtain stable task-level feedback while keeping per-candidate cost manageable.

To further control training overhead, early stopping is applied when validation improvement plateaus. This design yields comparable performance estimates across candidate architectures while avoiding excessive computation on clearly underperforming models. The resulting training output is summarized in a structured form, including the final validation metric, convergence status, and model checkpoint path, and is then passed to the subsequent evaluation and refinement stages.

By separating controlled training from deployment verification, AutoMCU avoids conflating statistical task performance with hardware feasibility. This distinction is particularly important in MCU scenarios, where the most accurate model is not necessarily deployable under strict memory and backend constraints.

\begin{figure*}[t]
  \centering
  \includegraphics[width=0.90\textwidth]{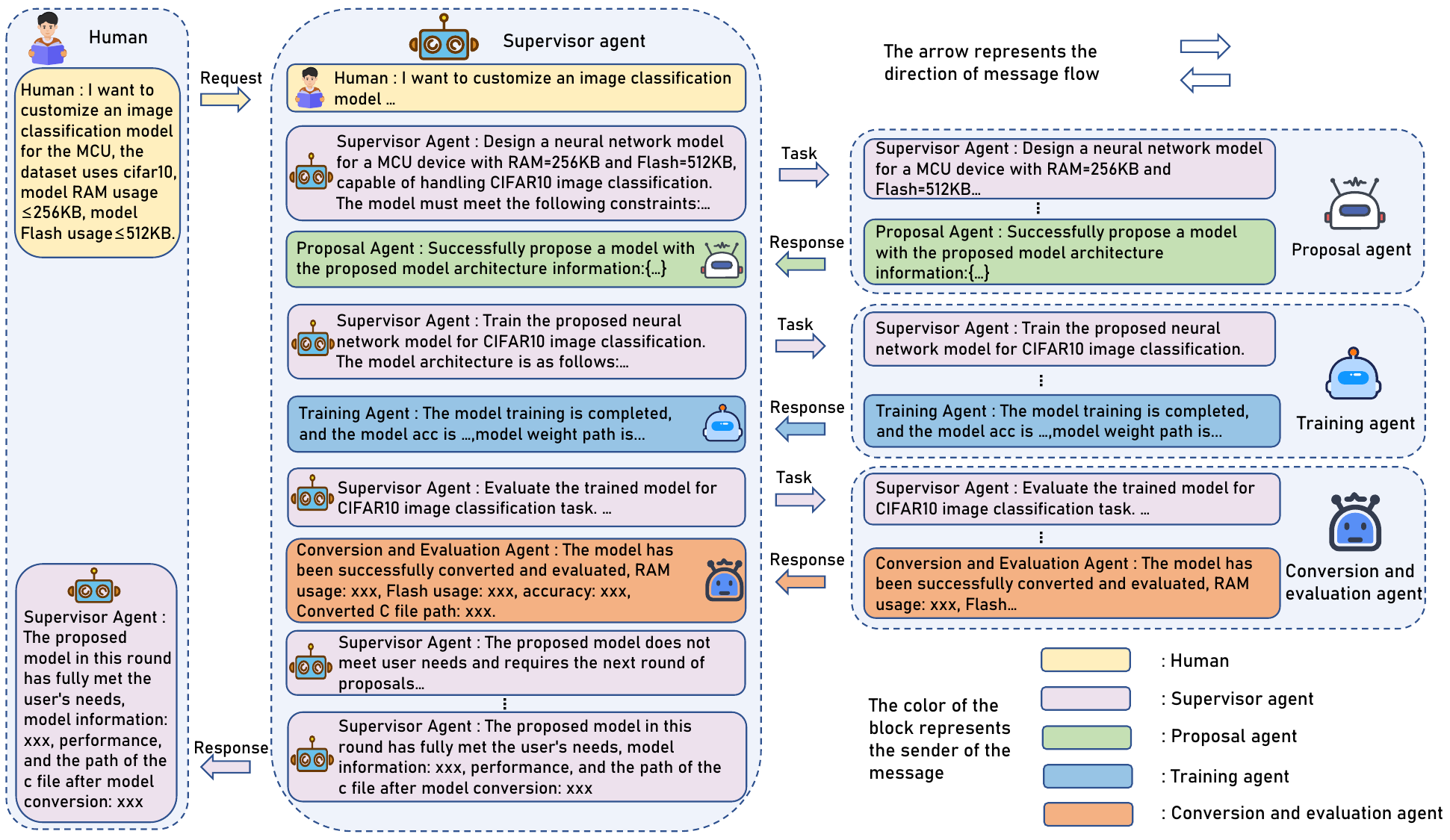}
  \caption{Multi-Agent Scheduling and Interaction Mechanism. The Supervisor Agent coordinates multiple specialized agents through structured messaging, where each agent executes tasks with isolated internal states and returns only summarized results, ensuring stable and controllable iterative optimization.}
  \label{Fig:collaboration}
\end{figure*}

\subsection{Backend-Verified Deployment Evaluation}
A common limitation of hardware-aware search is the gap between proxy efficiency metrics and actual deployment behavior. Models with favorable parameter counts or FLOPs may still fail on a target MCU because of memory allocation patterns, unsupported operators, or backend-specific conversion constraints. AutoMCU addresses this deployment gap by incorporating the target toolchain directly into the optimization loop.

After candidate construction, and again after training when needed for final validation, the evaluation and conversion module invokes the vendor backend to assess deployability on the target platform. In the current implementation, this process is based on STM32Cube.AI, which provides deployment-oriented measurements such as estimated SRAM and Flash usage, operator compatibility information, and deployment-ready C code generation for STM32 MCUs. Equivalent backends can be substituted for other platforms.

For a trained candidate architecture $a$, the backend returns deployment-related measurements
\begin{equation}
M(a;C) = \big(\texttt{RAM}(a), \texttt{Flash}(a), \texttt{status}(a)\big),
\label{eq:measure}
\end{equation}
where \texttt{RAM}$(a)$ and \texttt{Flash}$(a)$ denote backend-reported memory usage and \texttt{status}$(a)$ indicates whether deployment preparation succeeds under the target constraints. These measurements are treated as implementation-grounded deployment feedback rather than proxy estimates.

The Evaluation and Conversion Agent produces a deployment report containing task performance, RAM usage, Flash usage, operator compatibility status, and deployment artifact paths. This report serves two purposes: first, it determines whether the current model satisfies user-specified deployment constraints and can be accepted as the final output; second, it provides precise feedback for subsequent proposal refinement based on actual backend behavior rather than abstract analytical estimates.

By tightly integrating model evaluation with the actual deployment preparation process, AutoMCU closes the gap between architecture exploration and MCU deployment. This backend-grounded evaluation design enables the system to optimize toward models that are not only accurate but also concretely deployable on the target platform.

\subsection{Multi-Agent Scheduling and Interaction Mechanism}
AutoMCU adopts a supervisor-driven Multi-Agent Scheduling and Interaction Mechanism (MSIM) to coordinate the iterative model customization process. The Supervisor Agent serves as a centralized controller that interprets user requests, orchestrates task execution, and determines the termination of the optimization loop.

In this design, each task agent operates with isolated internal states and is exposed only to task-relevant information provided by the supervisor, as shown in Fig. \ref{Fig:collaboration}. Intermediate reasoning processes, tool invocation details, and internal decision traces of individual agents are not shared across agents. Instead, each agent returns a structured summary of its execution results, including task status and key results, to the supervisor. The supervisor then integrates these summaries to guide subsequent decisions, such as triggering further optimization, reassigning failed tasks, or terminating workflows after meeting user needs.

This interaction protocol contrasts with the shared-context multi-agent paradigm, where all agents operate on a shared and continuously growing message history. This shared-context design often leads to excessive context accumulation, increased token consumption, and dilution of task-critical information. Furthermore, mixing messages from multiple agents with different roles can introduce unintended policy disruptions and unstable behaviors in long-horizon LLM workflows, such as agents performing tasks outside their responsibilities.

By adopting centralized scheduling and strict state isolation, AutoMCU limits shared context to concise, task-level summaries, significantly reducing contextual redundancy and improving inference efficiency. At the same time, isolating agent state helps prevent cross-agent interference and enhances the overall system controllability and predictability.

This design is particularly suitable for neural network customization under MCU constraints, where iterative optimization under strict hardware budgets requires stability and efficiency. Supervisor-driven coordination mechanisms enable AutoMCU to scale to longer optimization horizons while maintaining reliable behavior and repeatable results.

\subsection{Closed-Loop Optimization}
AutoMCU operates as a closed-loop optimization framework that iteratively refines candidate architectures under explicit task and hardware constraints. The optimization follows a proposal--screen--train--evaluate cycle coordinated by the Supervisor Agent, with each stage producing structured feedback that constrains subsequent decisions.

Let $\mathcal{R}^{(k)}$ denote the historical evaluation repository accumulated up to iteration $k$. At each iteration, the Proposal Agent generates new candidate architectures conditioned on the current task requirements, hardware constraints, and selected summaries from $\mathcal{R}^{(k)}$. These candidates are first screened for feasibility through the deployment backend, then trained under the controlled protocol, and finally evaluated through backend-verified deployment analysis.

The repository is updated iteratively as
\begin{equation}
\mathcal{R}^{(k+1)} =
\mathcal{R}^{(k)} \cup
\left\{
\big(a, P(a;\mathcal{D},T), M(a;C)\big)
\right\},
\label{eq:update}
\end{equation}
for each evaluated candidate $a$ generated at iteration $k$, where $P(a;\mathcal{D},T)$ denotes task-level performance and $M(a;C)$ denotes backend-reported deployment measurements under constraint set $C$.

Hardware feasibility is enforced as a hard constraint throughout this loop. Candidate models that violate RAM or Flash budgets, or fail backend processing, are rejected before training, ensuring that optimization remains focused on the deployable region. The Proposal Agent then leverages structured historical feedback to incrementally refine architectures instead of repeatedly exploring clearly infeasible candidates.

The optimization loop terminates once user-defined requirements are satisfied or a maximum iteration budget is reached. Through explicit constraint enforcement, structured feedback, and backend-grounded evaluation, AutoMCU supports stable convergence toward deployable MCU models.

\section{Performance Evaluation}\label{sec:performance_evaluation}
\subsection{Experimental Setup}

\textbf{Environment Setting.} Our experiments were conducted on an Ubuntu 22.04 server featuring an AMD EPYC 7642 CPU and an NVIDIA RTX 3090 GPU with 24GB of VRAM. For the LLM component of AutoMCU, we employed the DeepSeek-V3.2\cite{liu2025deepseek} API as the core engine for architectural proposal generation.

\textbf{LLM Configuration.} AutoMCU is designed to support multiple large language model (LLM) backends and does not rely on a specific LLM implementation. 
While the quality of the underlying LLM may influence the efficiency of the optimization process and the final model performance, the AutoMCU framework itself remains applicable across different LLMs.
In our experiments, we primarily adopt DeepSeek-V3.2\cite{liu2025deepseek} as the default LLM backend for all main experiments, ablation studies, and other evaluations, in order to ensure experimental consistency and reproducibility. 
To evaluate the robustness of AutoMCU across different LLM backends, we additionally conduct experiments on CIFAR-10\cite{krizhevsky2009learning} and CIFAR-100\cite{krizhevsky2009learning} using Mimo-V2-Flash\cite{mimo2025flash} and Qwen-Plus\cite{qwen2.5}. These datasets are chosen as representative benchmarks due to their moderate complexity and widespread adoption.
For all LLMs, the sampling temperature is fixed to 0, eliminating stochasticity and ensuring deterministic behavior during the multi-agent optimization process.

\textbf{Baseline Models.} For a comprehensive performance evaluation, we adopt four types of baseline models, described as follows:
\begin{itemize}
\item \textbf{MCU-oriented HW-NAS baselines.}
We include \textit{$\mu$NAS}\cite{liberis2021munas} and \textit{ColabNAS}\cite{garavagno2024colabnas} as the two methods most closely related to our work. $\mu$NAS is a constrained neural architecture search framework specifically designed for microcontrollers, which explicitly models peak RAM usage, model size in Flash, and latency during search through a fine-grained search space and evolutionary optimization. ColabNAS is an affordable hardware-aware NAS method for generating lightweight task-specific CNNs under hardware constraints. It adopts a derivative-free search strategy inspired by Occam’s razor and progressively expands model complexity while checking resource feasibility. Since both methods explicitly target compact model search on resource-constrained devices, they form the most relevant HW-NAS baselines for AutoMCU.

\item \textbf{LLM-based NAS baseline.}
We further include \textit{GENIUS}\cite{zheng2023can} as a representative LLM-based NAS method. GENIUS formulates neural architecture search as an iterative language-guided optimization process, where a large language model proposes candidate architectures and refines them according to performance feedback from previously evaluated models. Unlike $\mu$NAS, ColabNAS, and AutoMCU, GENIUS is not designed for hardware-aware search or deployment-oriented customization under strict MCU constraints. In contrast, AutoMCU is developed for automated neural network customization under strict MCU constraints and integrates architecture proposal, feasibility screening, controlled training, and backend-verified deployment evaluation into a unified closed-loop workflow. Therefore, we compare AutoMCU with GENIUS on NAS-Bench-201, a standardized NAS benchmark, to fairly evaluate the architecture search capability of both methods without the influence of MCU-specific deployment factors.

\item \textbf{Open-source NAS-generated models}: MCUNet \cite{lin2020mcunet}, MnasNet \cite{tan2019mnasnet}, and FairNAS \cite{chu2021fairnas} are representative NAS-based lightweight models. Among them, MCUNet explicitly targets MCU deployment, while MnasNet and FairNAS provide strong references for efficient architecture design under resource-constrained settings.

\item \textbf{Classical lightweight models}: MobileNetV2\cite{sandler2018mobilenetv2}, SqueezeNet\cite{iandola2016squeezenet}, and ShuffleNetV2\cite{ma2018shufflenet} are widely used lightweight models known for their efficiency. These models are trained for 200 epochs under identical configurations to provide a fair comparison against the NAS-based and MCU-optimized approaches.
\end{itemize}

Since $\mu$NAS, ColabNAS, and AutoMCU are designed for model customization on resource-constrained MCUs, we evaluate them under two deployment settings: RAM$\leq$1024KB and Flash$\leq$1024KB, and RAM$\leq$256KB and Flash$\leq$512KB. For the other lightweight baselines, we report results under the relatively relaxed setting of RAM$\leq$1024KB and Flash$\leq$1024KB to assess whether they satisfy typical MCU deployment requirements.
\begin{table*}

\centering
\caption{Performance Evaluation.
\textcolor{green}{Green} indicates compliance with resource constraints, while \textcolor{red}{red} indicates exceeding resource constraints. Models are grouped by deployment constraints, including RAM$\leq$1024KB and Flash$\leq$1024KB, and RAM$\leq$256KB and Flash$\leq$512KB. Classical lightweight models and open-source NAS-generated models are evaluated under the RAM$\leq$1024KB and Flash$\leq$1024KB setting.}
\label{main_results}
\resizebox{\linewidth}{!}{
\begin{tabular}{l l *{5}{c} *{5}{c}}
\toprule
\multirow{3}{*}{\textbf{Constraints}} & 
\multirow{3}{*}{\textbf{Model}} & 
\multicolumn{5}{c}{\textbf{CIFAR10}} \\
\cmidrule(lr){3-7} 
& & Accuracy  & RAM & Flash & Time & Cost  \\
& & (\%) $\uparrow$ & (KB) $\downarrow$ & (KB) $\downarrow$ & (GPU Hours) $\downarrow$ & (\$) $\downarrow$ \\
\midrule
\multirow{10}{*}{\begin{tabular}[c]{@{}l@{}}RAM$\leq$1024KB\\Flash$\leq$1024KB\end{tabular}}
& MobileNetV2 \cite{sandler2018mobilenetv2} & 84.50 & {\cellcolor{green!30}110.44} & {\cellcolor{red!30}8670.41} & N/A & N/A \\
& ShuffleNetV2 \cite{ma2018shufflenet} & 79.34 & {\cellcolor{green!30}24.00} & {\cellcolor{red!30}4905.33} & N/A & N/A \\
& SqueezeNet \cite{iandola2016squeezenet} & 82.91 & {\cellcolor{green!30}72.00} & {\cellcolor{red!30}2892.79} & N/A & N/A \\
& MCUNet-in2 \cite{lin2020mcunet} & 83.69 & {\cellcolor{green!30}42.63} & {\cellcolor{red!30}2197.57} & N/A & N/A \\
& MnasNet \cite{tan2019mnasnet} & 83.61 & {\cellcolor{green!30}54.75} & {\cellcolor{red!30}12094.38} & N/A & N/A \\
& FairNAS-C \cite{chu2021fairnas} & 84.25 & {\cellcolor{green!30}56.44} & {\cellcolor{red!30}12159.85} & N/A & N/A \\
& $\mu$NAS \cite{liberis2021munas} & \textbf{89.75} & {\cellcolor{red!30}1159.99} & {\cellcolor{red!30}2902.15} & 257.48 & 53.35 \\
& ColabNAS \cite{garavagno2024colabnas} & 74.77 & {\cellcolor{green!30}163.00} & {\cellcolor{green!30}764.49}& 2.32 & 0.49 \\
& AutoMCU (DeepSeek-V3.2\cite{liu2025deepseek}) & \underline{89.31±0.16} & {\cellcolor{green!30}218.81±70.71 [135.00 , 364.31]} & {\cellcolor{green!30}848.97±99.16 [724.85 , 988.67]} & 1.58±1.07 & \underline{0.45±0.31} \\
& AutoMCU (MiMo-V2-Flash\cite{mimo2025flash}) & 89.30±0.34 & {\cellcolor{green!30}210.24±65.74 [128.00 , 384.00]} & {\cellcolor{green!30}876.00±126.11 [629.28 , 1015.82]} & \textbf{1.00±0.48} & \textbf{0.23±0.11} \\
& AutoMCU (Qwen-Plus\cite{qwen2.5}) & 88.77±0.46 & {\cellcolor{green!30}185.66±59.76 [83.38 , 257.12]} & {\cellcolor{green!30}888.31±98.52 [707.18 , 1006.38]} & \underline{1.54±1.04} & 0.47±0.30 \\

\midrule
\multirow{3}{*}{\begin{tabular}[c]{@{}l@{}}RAM$\leq$256KB\\Flash$\leq$512KB\end{tabular}}
& $\mu$NAS \cite{liberis2021munas}& 87.88 & {\cellcolor{red!30}580.92} & {\cellcolor{red!30}872.34} & 173.87 & 36.03 \\
& ColabNAS \cite{garavagno2024colabnas} & 56.71 & {\cellcolor{green!30}162.50} & {\cellcolor{green!30}98.35}& 1.58 & 0.33 \\
& AutoMCU (DeepSeek-V3.2\cite{liu2025deepseek}) & 87.62±0.52 & {\cellcolor{green!30}124.84±23.80 [83.38 , 160.00]} & {\cellcolor{green!30} 466.55±34.61 [402.77 , 504.52]} & 1.56±1.25 & 0.45±0.37 \\
\bottomrule

\toprule
\multirow{3}{*}{\textbf{Constraints}} & 
\multirow{3}{*}{\textbf{Model}} & 
\multicolumn{5}{c}{\textbf{CIFAR100}} \\
\cmidrule(lr){3-7}
& & Accuracy  & RAM & Flash & Time & Cost \\
& & (\%) $\uparrow$ & (KB) $\downarrow$ & (KB) $\downarrow$ & (GPU Hours) $\downarrow$ & (\$) \\
\midrule
\multirow{10}{*}{\begin{tabular}[c]{@{}l@{}}RAM$\leq$1024KB\\Flash$\leq$1024KB\end{tabular}}
& MobileNetV2 \cite{sandler2018mobilenetv2} & 50.70 & {\cellcolor{green!30}110.44} & {\cellcolor{red!30}9120.77} & N/A & N/A \\
& ShuffleNetV2 \cite{ma2018shufflenet} & 44.73 & {\cellcolor{green!30}24.00} & {\cellcolor{red!30}5265.68} & N/A & N/A \\
& SqueezeNet \cite{iandola2016squeezenet} & 42.82 & {\cellcolor{green!30}72.00} & {\cellcolor{red!30}3073.14} & N/A & N/A \\
& MCUNet-in2 \cite{lin2020mcunet} & 51.06 & {\cellcolor{green!30}42.63} & {\cellcolor{red!30}2254.17} & N/A & N/A \\
& MnasNet \cite{tan2019mnasnet} & 52.34 & {\cellcolor{green!30}54.75} & {\cellcolor{red!30}12544.73}& N/A & N/A \\
& FairNAS-C \cite{chu2021fairnas} & 45.75 & {\cellcolor{green!30}56.44} & {\cellcolor{red!30}12610.20}& N/A & N/A \\
& $\mu$NAS \cite{liberis2021munas} & 60.26 & {\cellcolor{red!30}1024.10} & {\cellcolor{red!30}2522.94}& 188.60 & 39.08 \\
& ColabNAS \cite{garavagno2024colabnas} & 42.96 & {\cellcolor{green!30}322.50} & {\cellcolor{green!30}414.21}& 2.52 & 0.53 \\
& AutoMCU (DeepSeek-V3.2\cite{liu2025deepseek}) & \underline{61.63±0.50} & {\cellcolor{green!30}157.35±49.24 [112.50 , 257.12]} & {\cellcolor{green!30}861.72±81.10 [674.17 , 958.78]} & \textbf{0.72±1.00} & \underline{0.25±0.31} \\
& AutoMCU (MiMo-V2-Flash\cite{mimo2025flash}) & \textbf{61.63±0.42} & {\cellcolor{green!30}183.13±65.27 [93.80 , 289.00]} & {\cellcolor{green!30}872.20±82.83 [728.46 , 1013.46]} & 1.28±1.35 & 0.32±0.34 \\
& AutoMCU (Qwen-Plus\cite{qwen2.5}) & 61.55±0.81 & {\cellcolor{green!30}201.52±43.28 [125.06 , 256.00]} & {\cellcolor{green!30}866.86 ± 116.07 [680.83 ,1008.65]} & \underline{0.83±0.56} & \textbf{0.24±0.15} \\

\midrule
\multirow{3}{*}{\begin{tabular}[c]{@{}l@{}}RAM$\leq$256KB\\Flash$\leq$512KB\end{tabular}}
& $\mu$NAS \cite{liberis2021munas}& 58.82 & {\cellcolor{red!30}349.00} & {\cellcolor{red!30}1448.15}& 160.80 & 33.32 \\
& ColabNAS \cite{garavagno2024colabnas} & 41.26 & {\cellcolor{green!30}163.00} & {\cellcolor{green!30}371.02}& 2.15 & 0.45 \\
& AutoMCU (DeepSeek-V3.2\cite{liu2025deepseek}) & 58.70±0.71 & {\cellcolor{green!30}138.21±57.10 [72.95 , 256.00]} & {\cellcolor{green!30}481.49±28.43 [430.25 , 509.80]} & 2.43±1.15 & 0.76±0.37 \\
\bottomrule

\end{tabular}
}
\end{table*}

\textbf{Evaluation Metrics.} 
To evaluate the performance of AutoMCU and the baseline models, we use the following metrics. For AutoMCU, the performance results are based on the mean and standard deviation reports of ten independent replicates. For the RAM and Flash resource metrics, the maximum and minimum values in ten additional experiments are also shown:

\begin{itemize}
    \item \textit{Accuracy}: Since the task is image classification, accuracy is used to assess model performance. It is computed by testing the model on a specific dataset and determining the percentage of correctly classified images.
    
    \item \textit{RAM and Flash Usage}: All hardware metrics, including RAM and Flash usage, are measured using the STM32CubeAI toolchain. This approach ensures accuracy and eliminates any potential estimation bias, providing a fair comparison among different algorithms.
    
    \item \textit{Search Time}: Search time is measured in GPU hours. This metric is reported for AutoMCU, $\mu$NAS, and ColabNAS, as these methods involve dynamic search processes. The other lightweight baselines are predesigned models and therefore do not have associated search time metrics.
    
    \item \textit{Search Cost}: This refers to the total cost incurred for renting GPU resources and the consumption of LLM tokens during model customization. Costs are based on the cloud platform's resource billing standards. GPU rental rates are \$0.21/hour. The token billing rates for each model are as follows: DeepSeek-V3.2\cite{liu2025deepseek} input is \$0.28/M and output is \$0.42/M. Mimo-V2-Flash\cite{mimo2025flash} input is \$0.10/M, output is \$0.30/M; Qwen-Plus\cite{qwen2.5} input is \$0.40/M, output is \$1.20/M.
\end{itemize}

\textbf{Datasets.} We utilize the standardized CIFAR-10\cite{krizhevsky2009learning} and CIFAR-100\cite{krizhevsky2009learning} image classification datasets for model training, preserving their native 32×32 resolution across all experiments to accurately reflect the constraints of MCUs deployments. This design choice is driven by critical RAM limitations in MCUs environments: while conventional approaches typically upsample images to 224×224 for enhanced accuracy, this preprocessing step requires 147KB of memory per input, consuming 14.3\%-57.4\% of the typical MCUs RAM budget (ranging from 256KB to 1MB). In contrast, processing the original 32×32 images demands only 3KB, thereby conserving valuable memory for model execution. For fairness, the same preprocessing and data augmentation techniques are applied across all datasets. Additionally, we incorporated the MNIST\cite{lecun1998gradient} and FashionMNIST\cite{xiao2017fashion} datasets to further evaluate the performance of AutoMCU.

\subsection{Comparative Evaluation on MCU-Oriented Baselines}

\textbf{Overall Comparison.} Table \ref{main_results} summarizes the comparison between AutoMCU and the MCU-oriented baselines under different resource constraints, together with several representative lightweight models.
As shown in Table\ref{main_results}, existing lightweight models and open-source NAS-generated models can often reduce RAM usage to some extent, but their Flash consumption still substantially exceeds the resource limits of typical MCUs. Even MCUNet, which exhibits relatively low RAM usage, still requires more than twice the Flash budget under the 1MB constraint. This imbalance between RAM and Flash usage makes it difficult for such models to simultaneously satisfy both memory and storage requirements in practical MCU deployment. Structurally, these models often rely on aggressive downsampling to reduce feature-map memory, but compensate for the resulting information loss by increasing channel width or network depth, which in turn sharply increases parameter count and Flash usage.

\textbf{Baseline Limitation Analysis.} Among the baselines, $\mu$NAS and ColabNAS are the most directly comparable methods to AutoMCU because both are designed for compact model search on resource-constrained devices. Nevertheless, their search behavior differs substantially from that of AutoMCU.
\textbf{a) $\mu$NAS} formulates MCU model design as a constrained multi-objective NAS problem and explicitly models peak RAM usage, model size, and latency. This makes it a strong MCU-oriented baseline. However, it relies on a fine-grained search space together with evolutionary optimization, which leads to high search cost and long customization time. As shown in Table\ref{main_results}, although $\mu$NAS achieves competitive and in some cases slightly higher accuracy, the search process is extremely time-consuming, requiring on average hundreds of GPU hours. Moreover, because it optimizes multiple objectives through weighted scalarization rather than strict hard filtering before training, the resulting architectures may still exceed the target deployment constraints.
\textbf{b) ColabNAS} adopts a more lightweight hardware-aware search strategy. Its derivative-free optimization procedure progressively increases model complexity according to an Occam’s-razor-inspired principle, which significantly reduces search overhead compared with conventional HW-NAS approaches. However, this progressive and relatively restricted search strategy also limits its ability to discover highly competitive architectures under more challenging datasets or strict dual constraints on RAM and Flash. As a result, although ColabNAS is efficient in search cost, its final accuracy is clearly lower than that of both $\mu$NAS and AutoMCU in our experiments.

\textbf{Effectiveness under MCU Constraints.} In contrast, AutoMCU consistently customizes resource-compliant models while maintaining strong accuracy under both relaxed and strict MCU constraints. Under the RAM$\leq$1024KB and Flash$\leq$1024KB setting, AutoMCU achieves performance that is competitive with $\mu$NAS while reducing customization time from hundreds of GPU hours to approximately 1--2 hours. Under the stricter RAM$\leq$256KB and Flash$\leq$512KB setting, AutoMCU continues to find feasible models with strong performance, whereas the advantages of traditional lightweight baselines further diminish. These results indicate that AutoMCU achieves a better balance among accuracy, deployment feasibility, and search efficiency than existing MCU-oriented baselines.
The efficiency advantage of AutoMCU can be attributed to its feasibility-first design. First, candidate architectures are screened against hardware constraints before training, which avoids wasting computation on clearly infeasible models. Second, the LLM-guided proposal mechanism leverages structured historical feedback to refine later candidates toward more promising regions of the search space. Compared with the more expensive evolutionary exploration used in $\mu$NAS and the progressively restricted heuristic exploration used in ColabNAS, this closed-loop refinement process leads to substantially faster convergence under strict deployment budgets.

\textbf{Low-resource Evaluation.} Furthermore, we evaluate AutoMCU under extremely resource-constrained settings, as shown in Table\ref{low_resource}. For example, under an ultra-low resource budget of RAM$\leq$64KB and Flash$\leq$64KB, AutoMCU is still able to automatically generate feasible models with average accuracies of 76.36\%, 99.23\%, and 91.28\% on CIFAR-10, MNIST, and FashionMNIST, respectively. These results further validate the practicality of AutoMCU under ultra-low-resource MCU constraints.

\begin{table*}[ht]
\centering
\caption{Customization of Low-Resource Models}
\label{low_resource}
\setlength{\tabcolsep}{3pt} 
\resizebox{\linewidth}{!}{
\begin{tabular}{c|ccccc}
\toprule
\makecell[c]{\textbf{Dataset}} &
\makecell[c]{\textbf{Constraints} \\ RAM/Flash} &
\makecell[c]{\textbf{Acc.} \\ (\%) $\uparrow$} &
\makecell[c]{\textbf{RAM} \\ (KB) $\downarrow$} &
\makecell[c]{\textbf{Flash} \\ (KB) $\downarrow$} &
\makecell[c]{\textbf{Time / Token} \\ (GPU Hours $\downarrow$ / M $\downarrow$)} \\
\midrule

\multirow{4}{*}{CIFAR10}
& 64/64   & 76.36±1.92 & 46.71±11.17 [26.89 , 63.50] & 49.27±11.01 [34.81 , 62.53] & 0.40±0.19 / 0.15±0.08 \\
& 64/128  & 79.03±1.65 & 49.58±7.64 [44.11 , 62.16] & 70.16±15.20 [52.74 , 98.85] & 0.30±0.15 / 0.11±0.07 \\
& 128/128 & 81.70±1.20 & 58.37±15.72 [44.11 , 96.00] & 97.48 ± 16.69 [65.02 , 124.30] & 0.47±0.27 / 0.20±0.17 \\
& 128/256 & 84.12±1.01 & 81.50±22.67 [52.11 , 128.00] & 173.81±45.24 [96.18 , 230.86] & 0.50±0.32 / 0.20±0.18 \\
\midrule

\multirow{4}{*}{MNIST}
& 64/64   & 99.23±0.13 & 48.98±14.78 [20.69 , 62.22] & 40.67±12.51 [18.59 , 59.45] & 0.36±0.25 / 0.12±0.08 \\
& 64/128  & 99.36±0.08 & 53.87±9.40 [39.85 , 62.22] & 55.88 ± 17.09 [36.46 , 91.98] & 0.36±0.19 / 0.12±0.06 \\
& 128/128 & 99.47±0.16 & 62.90±24.42 [24.08 , 98.00] & 110.66±12.96 [85.99 , 127.75] & 0.34±0.08 / 0.14±0.09 \\
& 128/256 & 99.48±0.11 & 87.50±26.10 [45.97 , 125.61] & 142.51±72.68 [40.68 , 253.52] & 0.30±0.05 / 0.09±0.02 \\
\midrule

\multirow{4}{*}{\makecell[c]{Fashion\\MNIST}}
& 64/64   & 91.28±0.41 & 53.23±12.96 [24.80 , 62.22] & 43.45±9.36 [26.69 , 58.50] & 0.28±0.04 / 0.11±0.04 \\
& 64/128  & 91.92±0.50 & 55.03±11.50 [27.60 , 62.22] & 65.92±16.41 [44.91 , 90.15] & 0.34±0.14 / 0.11±0.04 \\
& 128/128 & 92.26±0.48 & 74.27±29.21 [31.11 , 122.50] & 82.79±28.17 [45.20 , 127.75] & 0.41±0.18 / 0.13±0.05 \\
& 128/256 & 92.72±0.46 & 85.09±27.33 [27.60 , 122.50] & 192.13±49.80 [106.62 , 249.45] & 0.29±0.10 / 0.11±0.03 \\
\bottomrule

\end{tabular}
}
\end{table*}

\begin{table*}[t]
\centering
\caption{Comparison with GENIUS on NAS-Bench-201. Both methods use DeepSeek-V3.2 with temperature set to 0. For each dataset, the search is repeated 10 times, and each run iteratively searches 10 architectures. We report the mean and standard deviation of validation and test accuracy (\%).}
\label{nasbench201_genius}
\resizebox{\textwidth}{!}{
\begin{tabular}{l c cc cc cc}
\toprule
\multirow{2}{*}{\textbf{Method}} & \multirow{2}{*}{\textbf{Search (Archs)}} & \multicolumn{2}{c}{\textbf{CIFAR-10}} & \multicolumn{2}{c}{\textbf{CIFAR-100}} & \multicolumn{2}{c}{\textbf{ImageNet16-120}} \\
\cmidrule(lr){3-4} \cmidrule(lr){5-6} \cmidrule(lr){7-8}
 &  & Validation (\%) & Test (\%) & Validation (\%) & Test (\%) & Validation (\%) & Test (\%) \\
\midrule
GENIUS\cite{zheng2023can}  & 10 & 90.82 $\pm$ 0.84 & 93.73 $\pm$ 0.54 & 71.37 $\pm$ 1.14 & 71.30 $\pm$ 0.98 & 44.63 $\pm$ 2.28 & 44.51 $\pm$ 2.49 \\
AutoMCU & 10 & \textbf{90.89 $\pm$ 0.64} & \textbf{93.74 $\pm$ 0.40} & \textbf{72.32 $\pm$ 1.09} & \textbf{72.24 $\pm$ 1.21} & \textbf{45.07 $\pm$ 0.85} & \textbf{45.19 $\pm$ 0.95} \\
\bottomrule
\end{tabular}
}
\end{table*}

\begin{table*}
\centering
\caption{Ablation Study (CIFAR10)}
\label{Ablation}
\resizebox{\linewidth}{!}{
\begin{tabular}{ccccccccc}
\toprule
\multirow{2}{*}{\textbf{Model}}
& \textbf{Accuracy} & \textbf{RAM} & \textbf{Flash} & \textbf{Time} (GPU & \textbf{Token} & \textbf{Failure Rate} \\
& (\%) $\uparrow$ & (KB) $\downarrow$ & (KB) $\downarrow$ &  Hours) $\downarrow$ & (M) $\downarrow$ & (\%) $\downarrow$ \\
\midrule 
AutoMCU & 87.62±0.52 & 124.84±23.80 [83.38 , 160.00] & 466.55±34.61 [402.77 , 504.52] & 1.56±1.25 & 0.46±0.40 & 0\\
w/o MSIM  & 87.48±0.98 & 111.10±28.20 [80.00 , 144.63] & 463.03 ± 64.14 [354.65 , 511.95] & 1.28±1.04 & 0.99±1.13 & 50\\
w/o HAG & 82.89±2.81 & 108.51±53.43 [24.00 , 180.44] & 299.60 ± 100.12 [169.03 , 490.61] & 3.81±0.85 & 0.54±0.07 & 0\\
Baseline & 81.06±3.72 & 147.32±16.73 [128.00 , 168.84] & 228.50±166.61 [47.05 , 444.64] & 3.36±0.33 & 1.92±0.28 & 60\\
\bottomrule
\end{tabular}
}
\end{table*}

\subsection{Comparison with LLM-Based NAS on NAS-Bench-201}

\textbf{Evaluation Protocol.} To further evaluate the architecture search capability of AutoMCU against LLM-based NAS methods in a standardized and reproducible setting, we compare AutoMCU with GENIUS on NAS-Bench-201\cite{dong2020nasbench201}. This benchmark is adopted because GENIUS is not designed for MCU-oriented HW-NAS or deployment-constrained customization, whereas AutoMCU is primarily developed for automated model customization under MCU constraints. Therefore, using NAS-Bench-201 enables a fair comparison between the two methods by focusing on search effectiveness itself rather than MCU-specific deployment factors.
For fairness, both GENIUS and AutoMCU use the same LLM backend, namely DeepSeek-V3.2, with the sampling temperature fixed to 0. On each dataset, we repeat the search process 10 times independently, and each run is allowed to iteratively search 10 architectures. We conduct experiments on the three datasets provided by NAS-Bench-201, namely CIFAR-10, CIFAR-100, and ImageNet16-120. For each run, we select the best searched architecture according to validation accuracy and report both its validation and test accuracy. Final results are presented as mean and standard deviation over 10 repeated runs.

\textbf{Overall Performance.} Table \ref{nasbench201_genius} shows that AutoMCU consistently achieves better average performance than GENIUS on all three NAS-Bench-201 datasets under the same search budget of 10 architectures per run. The gain is relatively small on CIFAR-10, but becomes more evident on CIFAR-100 and ImageNet16-120, indicating that AutoMCU is more effective in handling more challenging search scenarios where the architecture-performance landscape becomes more complex.

\textbf{Search Stability.} A more important difference lies in search stability. AutoMCU exhibits lower variance than GENIUS on nearly all datasets, with the gap being especially large on ImageNet16-120. Specifically, the validation standard deviation is reduced from 2.28 in GENIUS to 0.85 in AutoMCU, and the test standard deviation is reduced from 2.49 to 0.95. This indicates that AutoMCU is not only slightly better on average, but also substantially more reliable across repeated runs. Such robustness is particularly important under a limited search budget, where a stable method is more likely to consistently identify strong architectures.

\textbf{Methodological Analysis.} The observed difference is closely related to the design philosophies of the two methods. GENIUS formulates NAS as a language-guided black-box optimization process, in which the LLM iteratively proposes an architecture, receives performance feedback, and generates the next candidate through prompting. This design demonstrates that a general-purpose LLM can indeed perform architecture search, but it also makes the optimization process strongly dependent on free-form generative reasoning. As a result, search trajectories may become unstable, especially on more difficult datasets where performance differences between candidate architectures are subtle and local optima are abundant.

\textbf{Structured Search Advantage.} By contrast, AutoMCU adopts a more structured search paradigm. Although it is originally developed for MCU-oriented automated model customization, its search loop explicitly organizes proposal generation, feedback summarization, and iterative refinement within a disciplined closed-loop workflow. Rather than relying solely on loosely structured textual exploration, AutoMCU uses constructible architecture representations and task-relevant historical feedback to steer subsequent proposals toward more promising regions of the search space. This more structured optimization process reduces ineffective exploration and improves consistency across runs.

\textbf{Summary.} These results highlight that the advantage of AutoMCU does not stem only from hardware-aware feasibility checking, but also from the overall design of its automated search workflow. Even on a standard NAS benchmark where MCU deployment constraints are removed, the proposed framework still achieves better average performance and substantially stronger robustness than a representative LLM-based NAS method. This suggests that workflow-level design is an important factor in making LLM-assisted architecture search practically effective.

\subsection{Ablation Studies}
To analyze the contribution of individual components in AutoMCU, we conduct a comprehensive ablation study on CIFAR-10 under strict MCU hardware constraints (RAM$\leq$256KB, Flash$\leq$512KB). Specifically, we construct multiple system variants by selectively removing key modules in AutoMCU:
\begin{itemize}
    \item \textbf{w/o MSIM}: removes the proposed Multi-Agent Scheduling and Interaction Mechanism and adopts a shared-context communication paradigm among agents;
    \item \textbf{w/o HAG}: removes the hardware-in-the-loop architecture generation mechanism and replaces the LLM-guided proposal process with random search;
    \item \textbf{Baseline}: removes both MSIM and HAG, resulting in a random-search-based workflow with shared agent context.
\end{itemize}

\textbf{Experimental Protocol.}
All methods are executed under the same hardware constraints and iteration budget. Each variant is independently evaluated over 10 runs, with the maximum number of iterations fixed to 10. For methods involving random search (w/o HAG and Baseline), each run generates up to 10 candidate architectures, and the most accurate feasible model is selected as the final result. For the w/o MSIM and Baseline variants, unstable inter-agent interactions may lead to execution failures such as invalid tool calls or inconsistent task assignment. Abnormally terminated runs are excluded from accuracy and resource statistics, while the failure rate is explicitly reported to characterize system robustness.

\textbf{Overall Results.}
Table\ref{Ablation} summarizes the ablation results in terms of model accuracy, RAM and Flash usage, customization time, token consumption, and failure rate. The full AutoMCU system achieves the best overall performance, obtaining the highest accuracy while maintaining low resource consumption and zero failure rate.

\textbf{Effect of MSIM.}
Comparing AutoMCU with w/o MSIM isolates the impact of the proposed Multi-Agent Scheduling and Interaction Mechanism. While w/o MSIM achieves comparable average accuracy on successful runs, it suffers from a 50\% failure rate, indicating severe instability during long-horizon multi-agent coordination. In addition, token consumption nearly doubles due to excessive context accumulation in the shared-history interaction pattern. These observations show that centralized supervision and state-isolated agent interaction are important not only for robustness, but also for controlling inference overhead.

\textbf{Effect of HAG.}
Removing HAG causes a notable degradation in search effectiveness. The average accuracy drops significantly, while both RAM and Flash usage show larger variance. Search time also increases substantially, indicating that random exploration struggles to efficiently identify high-quality architectures within a limited search budget. This confirms the importance of the structured proposal mechanism in guiding the search toward feasible and promising regions.

\textbf{Effect of Removing Both MSIM and HAG.}
The Baseline variant further amplifies these issues. It exhibits the lowest accuracy, the highest token consumption, and a 60\% failure rate, demonstrating that naive random exploration combined with shared-context multi-agent interaction is insufficient for reliable MCU model customization.

\textbf{Discussion.}
Overall, the ablation study confirms that both HAG and MSIM are indispensable components of AutoMCU. HAG improves search efficiency and model quality by enabling targeted architecture refinement under hardware constraints, while MSIM ensures stable and scalable multi-agent coordination throughout the optimization process. Notably, AutoMCU operates under a stricter feasibility-first regime than the random-search variants, since infeasible candidates are filtered early in the optimization loop. Despite this stricter setting, AutoMCU still consistently outperforms its ablated counterparts, demonstrating the effectiveness of the proposed design.

\begin{figure}[t]
  \centering
  \includegraphics[width=0.45\textwidth]{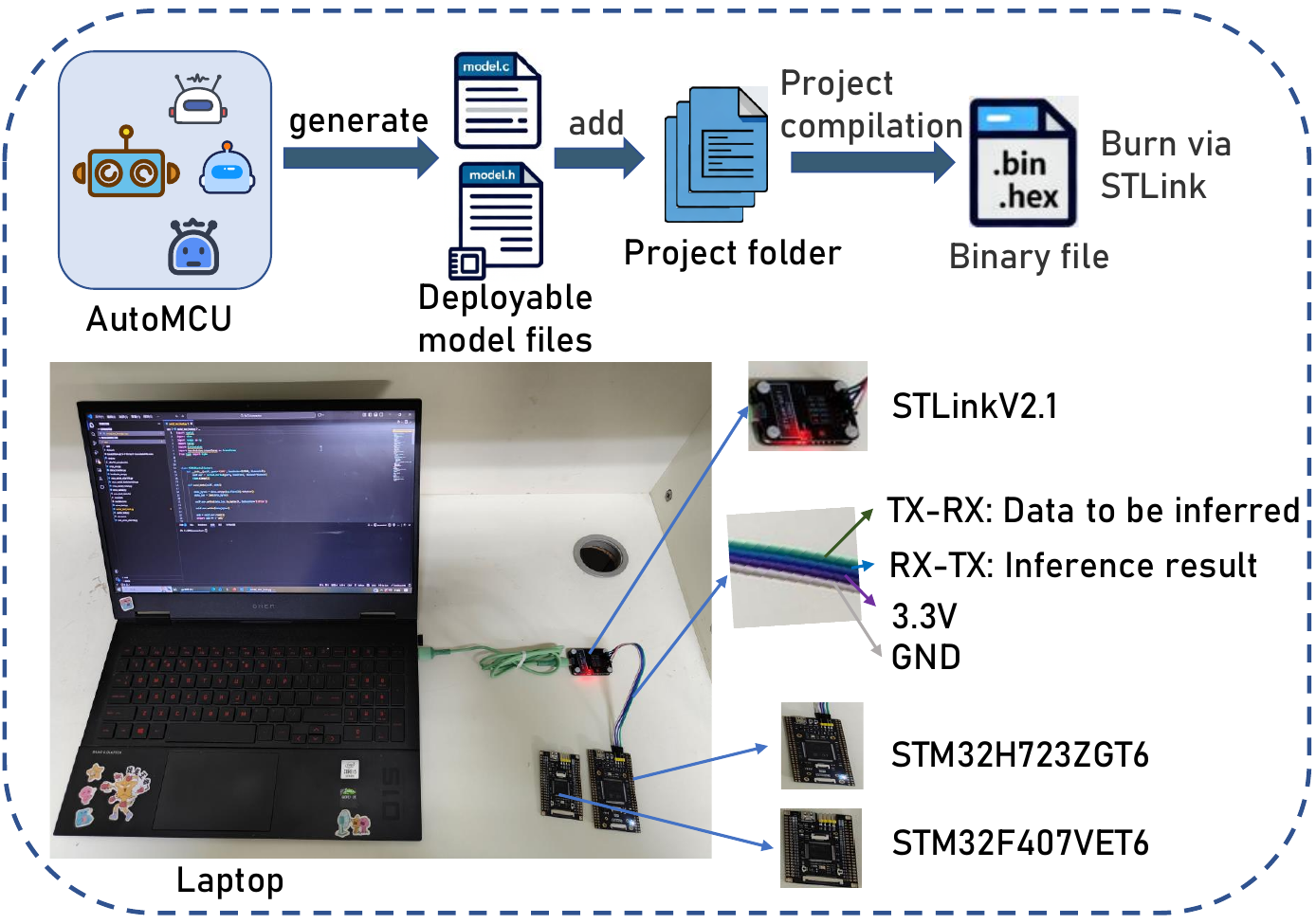}
  \caption{Workflow of model deployment and on-device inference validation. AutoMCU generates deployment-ready C code, which is then compiled into firmware using vendor toolchains and flashed onto target MCU devices for on-device inference testing.}
  \label{Fig:hardware}
\end{figure}

\subsection{Real-World Deployment and Adaptability Evaluation}
To evaluate the practical deployability and hardware adaptability of models customized by AutoMCU, we conduct real-world experiments on two commercial STM32 microcontroller platforms with distinct hardware specifications: STM32F407VET6 and STM32H723ZGT6 as shown in the Fig. \ref{Fig:hardware}. These two devices are deliberately selected to represent different levels of computational and memory resources, allowing us to verify AutoMCU’s capability to automatically tailor neural network architectures to heterogeneous MCU constraints.

The STM32F407VET6 features an ARM Cortex-M4 core operating at 168 MHz with 192 KB RAM and 512 KB Flash, representing a severely resource-constrained MCU. In contrast, the STM32H723ZGT6 is equipped with an ARM Cortex-M7 core running at 550 MHz, offering 564 KB RAM and 1 MB Flash, which enables higher-capacity models. This contrast allows us to examine whether AutoMCU can appropriately scale model capacity according to available hardware resources while preserving deployment feasibility.

\begin{table}[t]
\centering
\caption{Hardware experimentation.}
\label{Hardware experimentation}
\setlength{\tabcolsep}{3pt}
\begin{tabular}{l *{3}{c} *{3}{c}}
\toprule
\multirow{3}{*}{\textbf{Dataset}} & 
\multicolumn{3}{c}{\textbf{STM32F407VET6}} & 
\multicolumn{3}{c}{\textbf{STM32H723ZGT6}} \\
\cmidrule(lr){2-4} \cmidrule(lr){5-7}
& Accuracy  & RAM & Flash & Accuracy & RAM & Flash\\
& (\%)  & (KB)  & (KB) $\downarrow$ & (\%) $\uparrow$ & (KB) $\downarrow$ & (KB) $\downarrow$ \\
\midrule
MNIST & 99.36 & 39.85 & 97.68 & 99.37 & 73.50 & 139.80 \\
FashionMNIST & 93.42 & 66.13 & 423.70 & 93.94 & 82.66 & 653.86 \\
CIFAR10 & 86.39 & 72.95 & 312.04 & 88.99 & 114.19 & 922.84 \\
CIFAR100 & 58.05 & 83.38 & 437.22 & 61.66 & 125.06 & 813.72 \\
\bottomrule
\end{tabular}
\end{table}

For each platform, AutoMCU is applied to four vision benchmarks of increasing complexity: MNIST, Fashion-MNIST, CIFAR-10, and CIFAR-100. Given a specific dataset and target MCU, AutoMCU automatically customizes the network architecture under the corresponding hardware constraints. The system outputs deployable C source files generated via the vendor toolchain. The final firmware integration, flashing, and on-device execution are performed manually to validate real hardware feasibility.

All reported results correspond to models that are successfully compiled and executed on physical devices. Table \ref{Hardware experimentation} summarizes the on-device accuracy and memory footprint of the customized models. Across both platforms, AutoMCU consistently generates models that satisfy strict RAM and Flash constraints while achieving competitive accuracy across all datasets.

As expected, the more capable STM32H723ZGT6 allows AutoMCU to allocate greater model capacity, resulting in improved accuracy on more complex tasks such as CIFAR-10 and CIFAR-100. Meanwhile, under the severe constraints of the STM32F407VET6, AutoMCU still produces compact yet effective models, with accuracies of 86.39\% on CIFAR-10 and 58.05\% on CIFAR-100, demonstrating its robustness in extremely resource-limited scenarios.

These results confirm that AutoMCU can automatically adapt neural architectures under diverse MCU hardware constraints, effectively balancing accuracy and resource usage without manual intervention in the model design process.

\section{Conclusion and Future Work}\label{sec:Conclusion}

This paper presented AutoMCU, a feasibility-first LLM-based multi-agent system for end-to-end neural network customization under microcontroller constraints. By integrating real vendor deployment toolchains into a closed-loop optimization workflow, AutoMCU enables architecture proposal, feasibility screening, controlled training, and deployment verification within a unified framework. To support efficient and stable optimization, we introduced a hardware-in-the-loop architecture generation mechanism to eliminate infeasible candidates before training and a state-isolated multi-agent scheduling mechanism to improve long-horizon robustness. Experimental results under strict MCU constraints show that AutoMCU achieves competitive accuracy while reducing customization time from hundreds of GPU hours to approximately 1--2 hours compared with representative MCU-oriented HW-NAS baselines. Additional comparisons and real-device deployments further validate its architecture search capability, robustness, and practical applicability. 

Future work will extend AutoMCU beyond image classification to broader TinyML workloads, integrate more heterogeneous deployment backends and open-source embedded inference frameworks, and combine the proposed feasibility-first closed-loop design with more advanced architecture search and hardware modeling techniques to further improve customization quality, efficiency, and cross-platform adaptability.




\bibliographystyle{IEEEtran}
\bibliography{ref}

@inproceedings{garcia2021lspnet,
  title={Lspnet: A 2d localization-oriented spacecraft pose estimation neural network},
  author={Garcia, Albert and Musallam, Mohamed Adel and Gaudilliere, Vincent and Ghorbel, Enjie and Al Ismaeil, Kassem and Perez, Marcos and Aouada, Djamila},
  booktitle={Proceedings of the IEEE/CVF Conference on Computer Vision and Pattern Recognition},
  pages={2048--2056},
  year={2021}
}

@inproceedings{he2016deep,
  title={Deep residual learning for image recognition},
  author={He, Kaiming and Zhang, Xiangyu and Ren, Shaoqing and Sun, Jian},
  booktitle={Proceedings of the IEEE/CVF Conference on Computer Vision and Pattern Recognition},
  pages={770--778},
  year={2016}
}

@article{bahdanau2014neural,
  title={Neural machine translation by jointly learning to align and translate},
  author={Bahdanau, Dzmitry and Cho, Kyunghyun and Bengio, Yoshua},
  journal={arXiv preprint arXiv:1409.0473},
  year={2014}
}

@inproceedings{chan2016listen,
  title={Listen, attend and spell: A neural network for large vocabulary conversational speech recognition},
  author={Chan, William and Jaitly, Navdeep and Le, Quoc and Vinyals, Oriol},
  booktitle={IEEE International Conference on Acoustics, Speech and Signal Processing (ICASSP)},
  pages={4960--4964},
  year={2016},
  organization={IEEE}
}

@article{chorowski2015attention,
  title={Attention-based models for speech recognition},
  author={Chorowski, Jan K and Bahdanau, Dzmitry and Serdyuk, Dmitriy and Cho, Kyunghyun and Bengio, Yoshua},
  journal={Advances in Neural Information Processing Systems},
  volume={28},
  year={2015}
}

@inproceedings{pham2018efficient,
  title={Efficient neural architecture search via parameters sharing},
  author={Pham, Hieu and Guan, Melody and Zoph, Barret and Le, Quoc and Dean, Jeff},
  booktitle={International Conference on Machine Learning},
  pages={4095--4104},
  year={2018},
  organization={PMLR}
}

@inproceedings{sinha2021evolving,
  title={Evolving neural architecture using one shot model},
  author={Sinha, Nilotpal and Chen, Kuan-Wen},
  booktitle={Proceedings of the Genetic and Evolutionary Computation Conference},
  pages={910--918},
  year={2021}
}

@article{liu2018darts,
  title={Darts: Differentiable architecture search},
  author={Liu, Hanxiao and Simonyan, Karen and Yang, Yiming},
  journal={arXiv preprint arXiv:1806.09055},
  year={2018}
}

@article{xu2019pc,
  title={Pc-darts: Partial channel connections for memory-efficient architecture search},
  author={Xu, Yuhui and Xie, Lingxi and Zhang, Xiaopeng and Chen, Xin and Qi, Guo-Jun and Tian, Qi and Xiong, Hongkai},
  journal={arXiv preprint arXiv:1907.05737},
  year={2019}
}

@article{li2024zero,
  title={Zero-shot neural architecture search: Challenges, solutions, and opportunities},
  author={Li, Guihong and Hoang, Duc and Bhardwaj, Kartikeya and Lin, Ming and Wang, Zhangyang and Marculescu, Radu},
  journal={IEEE Transactions on Pattern Analysis and Machine Intelligence},
  volume={46},
  number={12},
  pages={7618--7635},
  year={2024},
  publisher={IEEE}
}

@inproceedings{lopes2021epe,
  title={Epe-nas: Efficient performance estimation without training for neural architecture search},
  author={Lopes, Vasco and Alirezazadeh, Saeid and Alexandre, Lu{\'\i}s A},
  booktitle={International Conference on Artificial Neural Networks},
  pages={552--563},
  year={2021},
  organization={Springer}
}

@inproceedings{lin2021zen,
  title={Zen-nas: A zero-shot nas for high-performance image recognition},
  author={Lin, Ming and Wang, Pichao and Sun, Zhenhong and Chen, Hesen and Sun, Xiuyu and Qian, Qi and Li, Hao and Jin, Rong},
  booktitle={Proceedings of the IEEE/CVF International Conference on Computer Vision},
  pages={347--356},
  year={2021}
}

@inproceedings{zoph2018learning,
  title={Learning transferable architectures for scalable image recognition},
  author={Zoph, Barret and Vasudevan, Vijay and Shlens, Jonathon and Le, Quoc V},
  booktitle={Proceedings of the IEEE/CVF Conference on Computer Vision and Pattern Recognition},
  pages={8697--8710},
  year={2018}
}

@inproceedings{wu2019fbnet,
  title={Fbnet: Hardware-aware efficient convnet design via differentiable neural architecture search},
  author={Wu, Bichen and Dai, Xiaoliang and Zhang, Peizhao and Wang, Yanghan and Sun, Fei and Wu, Yiming and Tian, Yuandong and Vajda, Peter and Jia, Yangqing and Keutzer, Kurt},
  booktitle={Proceedings of the IEEE/CVF Conference on Computer Vision and Pattern Recognition},
  pages={10734--10742},
  year={2019}
}

@inproceedings{dai2019chamnet,
  title={Chamnet: Towards efficient network design through platform-aware model adaptation},
  author={Dai, Xiaoliang and Zhang, Peizhao and Wu, Bichen and Yin, Hongxu and Sun, Fei and Wang, Yanghan and Dukhan, Marat and Hu, Yunqing and Wu, Yiming and Jia, Yangqing and others},
  booktitle={Proceedings of the IEEE/CVF Conference on Computer Vision and Pattern Recognition},
  pages={11398--11407},
  year={2019}
}

@inproceedings{tan2019mnasnet,
  title={Mnasnet: Platform-aware neural architecture search for mobile},
  author={Tan, Mingxing and Chen, Bo and Pang, Ruoming and Vasudevan, Vijay and Sandler, Mark and Howard, Andrew and Le, Quoc V},
  booktitle={Proceedings of the IEEE/CVF Conference on Computer Vision and Pattern Recognition},
  pages={2820--2828},
  year={2019}
}

@article{cai2018proxylessnas,
  title={Proxylessnas: Direct neural architecture search on target task and hardware},
  author={Cai, Han and Zhu, Ligeng and Han, Song},
  journal={arXiv preprint arXiv:1812.00332},
  year={2018}
}

@inproceedings{liberis2021munas,
  title={$\mu$nas: Constrained neural architecture search for microcontrollers},
  author={Liberis, Edgar and Dudziak, {\L}ukasz and Lane, Nicholas D},
  booktitle={Proceedings of the 1st Workshop on Machine Learning and Systems},
  pages={70--79},
  year={2021}
}

@article{lin2020mcunet,
  title={Mcunet: Tiny deep learning on iot devices},
  author={Lin, Ji and Chen, Wei-Ming and Lin, Yujun and Gan, Chuang and Han, Song and others},
  journal={Advances in Neural Information Processing Systems},
  volume={33},
  pages={11711--11722},
  year={2020}
}

@inproceedings{chu2021fairnas,
  title={Fairnas: Rethinking evaluation fairness of weight sharing neural architecture search},
  author={Chu, Xiangxiang and Zhang, Bo and Xu, Ruijun},
  booktitle={Proceedings of the IEEE/CVF International Conference on Computer Vision},
  pages={12239--12248},
  year={2021}
}

@inproceedings{sandler2018mobilenetv2,
  title={Mobilenetv2: Inverted residuals and linear bottlenecks},
  author={Sandler, Mark and Howard, Andrew and Zhu, Menglong and Zhmoginov, Andrey and Chen, Liang-Chieh},
  booktitle={Proceedings of the IEEE/CVF Conference on Computer Vision and Pattern Recognition},
  pages={4510--4520},
  year={2018}
}

@article{iandola2016squeezenet,
  title={SqueezeNet: AlexNet-level accuracy with 50x fewer parameters and< 0.5 MB model size},
  author={Iandola, Forrest N and Han, Song and Moskewicz, Matthew W and Ashraf, Khalid and Dally, William J and Keutzer, Kurt},
  journal={arXiv preprint arXiv:1602.07360},
  year={2016}
}

@inproceedings{ma2018shufflenet,
  title={Shufflenet v2: Practical guidelines for efficient cnn architecture design},
  author={Ma, Ningning and Zhang, Xiangyu and Zheng, Hai-Tao and Sun, Jian},
  booktitle={Proceedings of the European Conference on Computer Vision (ECCV)},
  pages={116--131},
  year={2018}
}

@techreport{krizhevsky2009learning,
  title        = {Learning Multiple Layers of Features from Tiny Images},
  author       = {Krizhevsky, Alex},
  institution  = {University of Toronto},
  year         = {2009},
  type         = {Technical Report}
}

@article{xiao2017fashion,
  title={Fashion-mnist: a novel image dataset for benchmarking machine learning algorithms},
  author={Xiao, Han and Rasul, Kashif and Vollgraf, Roland},
  journal={arXiv preprint arXiv:1708.07747},
  year={2017}
}

@article{lecun1998gradient,
author = {Lecun, Yann and Bottou, Leon and Bengio, Y. and Haffner, Patrick},
year = {1998},
month = {12},
pages = {2278 - 2324},
title = {Gradient-Based Learning Applied to Document Recognition},
volume = {86},
journal = {Proceedings of the IEEE},
doi = {10.1109/5.726791}
}

@article{han2015learning,
  title={Learning both weights and connections for efficient neural network},
  author={Han, Song and Pool, Jeff and Tran, John and Dally, William},
  journal={Advances in Neural Information Processing Systems},
  volume={28},
  year={2015}
}

@inproceedings{jacob2018quantization,
  title={Quantization and training of neural networks for efficient integer-arithmetic-only inference},
  author={Jacob, Benoit and Kligys, Skirmantas and Chen, Bo and Zhu, Menglong and Tang, Matthew and Howard, Andrew and Adam, Hartwig and Kalenichenko, Dmitry},
  booktitle={Proceedings of the IEEE Conference on Computer Vision and Pattern Recognition},
  pages={2704--2713},
  year={2018}
}

@article{deng2020model,
  title={Model compression and hardware acceleration for neural networks: A comprehensive survey},
  author={Deng, Lei and Li, Guoqi and Han, Song and Shi, Luping and Xie, Yuan},
  journal={Proceedings of the IEEE},
  volume={108},
  number={4},
  pages={485--532},
  year={2020},
  publisher={IEEE}
}

@article{he2021automl,
  title={AutoML: A survey of the state-of-the-art},
  author={He, Xin and Zhao, Kaiyong and Chu, Xiaowen},
  journal={Knowledge-based Systems},
  volume={212},
  pages={106622},
  year={2021},
  publisher={Elsevier}
}

@article{liang2023collaborative,
  title={A collaborative compression scheme for fast activity recognition on mobile devices via global compression ratio decision},
  author={Liang, Junjie and Zhang, Lei and Han, Chaolei and Bu, Can and Wu, Hao and Song, Aiguo},
  journal={IEEE Transactions on Mobile Computing},
  volume={23},
  number={4},
  pages={3259--3273},
  year={2023},
  publisher={IEEE}
}

@ARTICLE{11049020,
  author={Xie, Jianhang and Ding, Chuntao and Li, Xiaqing and Ren, Shenyuan and Li, Yidong and Lu, Zhichao},
  journal={IEEE Transactions on Mobile Computing}, 
  title={NestQuant: Post-Training Integer-Nesting Quantization for On-Device DNN}, 
  year={2025},
  volume={24},
  number={11},
  pages={12088-12102},
  publisher={IEEE}
}

@article{liu2020adadeep,
  title={AdaDeep: A usage-driven, automated deep model compression framework for enabling ubiquitous intelligent mobiles},
  author={Liu, Sicong and Du, Junzhao and Nan, Kaiming and Zhou, Zimu and Liu, Hui and Wang, Zhangyang and Lin, Yingyan},
  journal={IEEE Transactions on Mobile Computing},
  volume={20},
  number={12},
  pages={3282--3297},
  year={2020},
  publisher={IEEE}
}

@article{cha2025target,
  title={Target-Aware Neural Network Execution via Compiler-Guided Pruning},
  author={Cha, JooHyoung and Kim, Taeho and Lee, Jemin and Ha, Sangtae and Kwon, Yongin},
  journal={IEEE Transactions on Mobile Computing},
  year={2025},
  publisher={IEEE}
}

@inproceedings{
  cai2020once,
  title={Once for All: Train One Network and Specialize it for Efficient Deployment},
  author={Han Cai and Chuang Gan and Tianzhe Wang and Zhekai Zhang and Song Han},
  booktitle={International Conference on Learning Representations},
  year={2020},
  url={https://arxiv.org/pdf/1908.09791.pdf}
}

@article{david2021tensorflow,
  title={Tensorflow lite micro: Embedded machine learning for tinyml systems},
  author={David, Robert and Duke, Jared and Jain, Advait and Janapa Reddi, Vijay and Jeffries, Nat and Li, Jian and Kreeger, Nick and Nappier, Ian and Natraj, Meghna and Wang, Tiezhen and others},
  journal={Proceedings of Machine Learning and Systems},
  volume={3},
  pages={800--811},
  year={2021}
}

@misc{mimo2025flash,
  title={MiMo-V2-Flash Technical Report},
  author={LLM-Core Xiaomi},
  year={2025},
  url={https://github.com/XiaomiMiMo/MiMo-V2-Flash/blob/main/paper.pdf}
}

@article{liu2025deepseek,
  title={Deepseek-v3. 2: Pushing the frontier of open large language models},
  author={Liu, Aixin and Mei, Aoxue and Lin, Bangcai and Xue, Bing and Wang, Bingxuan and Xu, Bingzheng and Wu, Bochao and Zhang, Bowei and Lin, Chaofan and Dong, Chen and others},
  journal={arXiv preprint arXiv:2512.02556},
  year={2025}
}

@article{qwen2.5,
    title   = {Qwen2.5 Technical Report}, 
    author  = {An Yang and Baosong Yang and Beichen Zhang and Binyuan Hui and Bo Zheng and Bowen Yu and Chengyuan Li and Dayiheng Liu and Fei Huang and Haoran Wei and Huan Lin and Jian Yang and Jianhong Tu and Jianwei Zhang and Jianxin Yang and Jiaxi Yang and Jingren Zhou and Junyang Lin and Kai Dang and Keming Lu and Keqin Bao and Kexin Yang and Le Yu and Mei Li and Mingfeng Xue and Pei Zhang and Qin Zhu and Rui Men and Runji Lin and Tianhao Li and Tingyu Xia and Xingzhang Ren and Xuancheng Ren and Yang Fan and Yang Su and Yichang Zhang and Yu Wan and Yuqiong Liu and Zeyu Cui and Zhenru Zhang and Zihan Qiu},
    journal = {arXiv preprint arXiv:2412.15115},
    year    = {2024}
}

@inproceedings{guo2020single,
  title={Single path one-shot neural architecture search with uniform sampling},
  author={Guo, Zichao and Zhang, Xiangyu and Mu, Haoyuan and Heng, Wen and Liu, Zechun and Wei, Yichen and Sun, Jian},
  booktitle={European Conference on Computer Vision},
  pages={544--560},
  year={2020},
  organization={Springer}
}

@article{han2015deep,
  title={Deep compression: Compressing deep neural networks with pruning, trained quantization and huffman coding},
  author={Han, Song and Mao, Huizi and Dally, William J},
  journal={arXiv preprint arXiv:1510.00149},
  year={2015}
}

@inproceedings{wang2019haq,
  title={Haq: Hardware-aware automated quantization with mixed precision},
  author={Wang, Kuan and Liu, Zhijian and Lin, Yujun and Lin, Ji and Han, Song},
  booktitle={Proceedings of the IEEE/CVF Conference on Computer Vision and Pattern Recognition},
  pages={8612--8620},
  year={2019}
}

@inproceedings{zoph2017neural,
  title={Neural Architecture Search with Reinforcement Learning},
  author={Zoph, Barret and Le, Quoc},
  booktitle={International Conference on Learning Representations},
  year={2017}
}

@inproceedings{real2017large,
  title={Large-scale evolution of image classifiers},
  author={Real, Esteban and Moore, Sherry and Selle, Andrew and Saxena, Saurabh and Suematsu, Yutaka Leon and Tan, Jie and Le, Quoc V and Kurakin, Alexey},
  booktitle={International Conference on Machine Learning},
  pages={2902--2911},
  year={2017},
  organization={PMLR}
}

@inproceedings{luo2017thinet,
  title={Thinet: A filter level pruning method for deep neural network compression},
  author={Luo, Jian-Hao and Wu, Jianxin and Lin, Weiyao},
  booktitle={Proceedings of the IEEE international Conference on Computer Vision},
  pages={5058--5066},
  year={2017}
}

@inproceedings{he2017channel,
  title={Channel pruning for accelerating very deep neural networks},
  author={He, Yihui and Zhang, Xiangyu and Sun, Jian},
  booktitle={Proceedings of the IEEE International Conference on Computer Vision},
  pages={1389--1397},
  year={2017}
}

@article{garavagno2024colabnas,
  title={ColabNAS: Obtaining lightweight task-specific convolutional neural networks following Occam’s razor},
  author={Garavagno, Andrea Mattia and Leonardis, Daniele and Frisoli, Antonio},
  journal={Future Generation Computer Systems},
  volume={152},
  pages={152--159},
  year={2024},
  publisher={Elsevier}
}

@article{zheng2023can,
  title={Can gpt-4 perform neural architecture search?},
  author={Zheng, Mingkai and Su, Xiu and You, Shan and Wang, Fei and Qian, Chen and Xu, Chang and Albanie, Samuel},
  journal={arXiv preprint arXiv:2304.10970},
  year={2023}
}

@inproceedings{dong2020nasbench201,
  title     = {NAS-Bench-201: Extending the Scope of Reproducible Neural Architecture Search},
  author    = {Dong, Xuanyi and Yang, Yi},
  booktitle = {International Conference on Learning Representations (ICLR)},
  url       = {https://openreview.net/forum?id=HJxyZkBKDr},
  year      = {2020}
}

\begin{IEEEbiography}[{\includegraphics[width=1\textwidth]{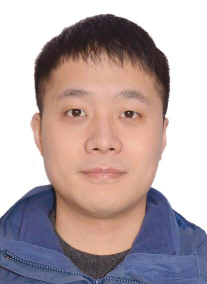}}]{Penglin Dai}(S'15-M'17) received the B.S. degree in mathematics and applied mathematics and the Ph.D. degree in computer science from Chongqing University, Chongqing, China, in 2012 and 2017, respectively. He is currently an Associate Professor with the School of Computing and Artificial Intelligence, Southwest Jiaotong University, Chengdu, China. His research interests include internet of vehicles, wireless networks and mobile computing.
\end{IEEEbiography}

\begin{IEEEbiography}[{\includegraphics[width=1\textwidth]{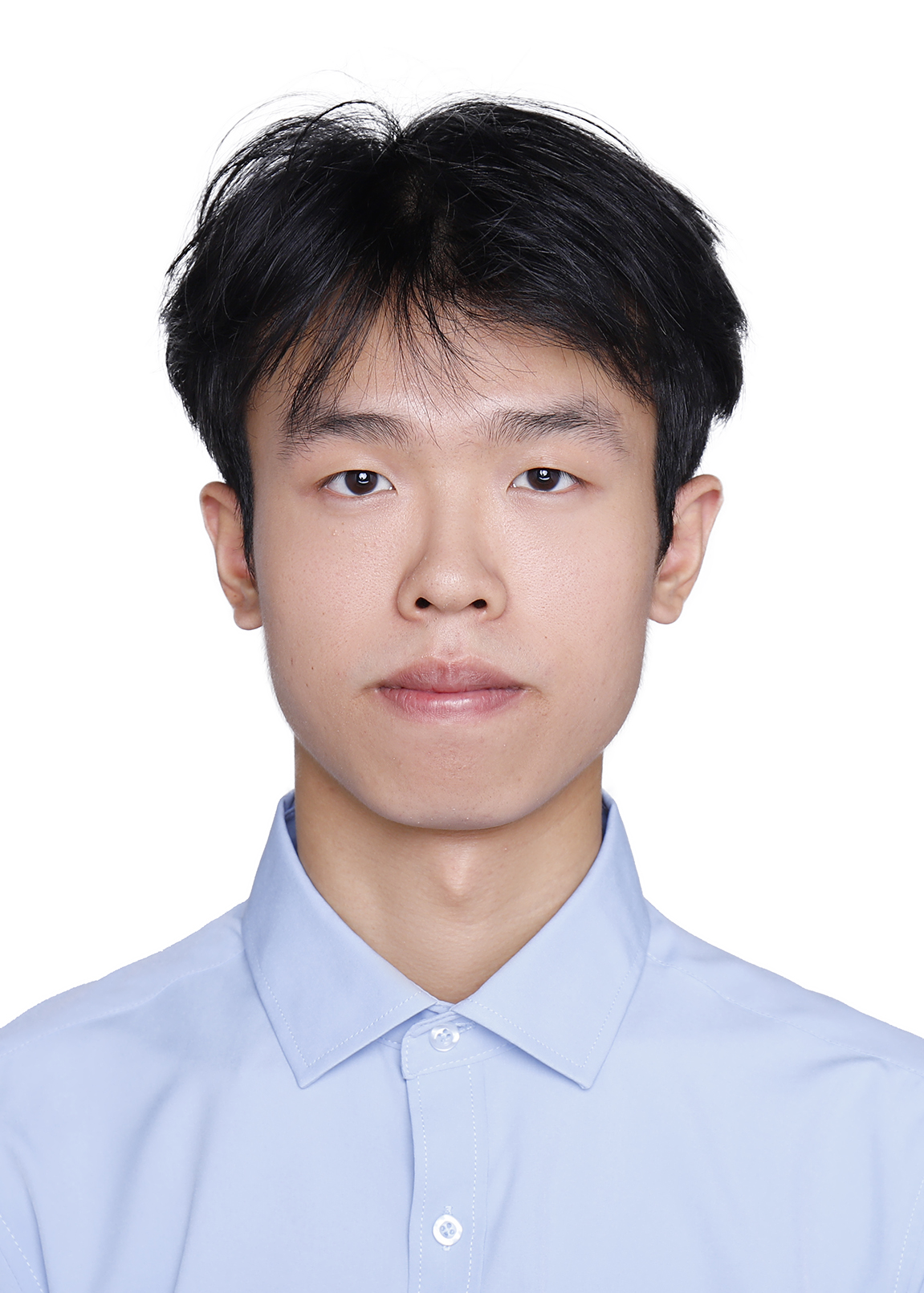}}]{Zijie Zhou} received the B.E. degree in artificial intelligence from the School of Computer and Artificial Intelligence, Southwest Jiaotong University, Chengdu, China. He is currently pursuing the master’s degree with the School of Computer and Artificial Intelligence, Southwest Jiaotong University. His research interests include AI agents and embedded artificial intelligence.
\end{IEEEbiography}

\begin{IEEEbiography}[{\includegraphics[width=1\textwidth]{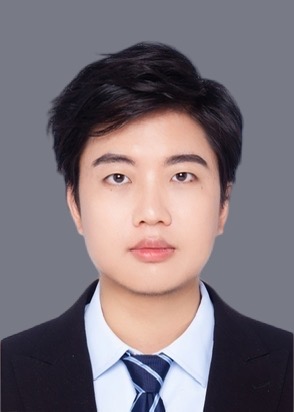}}]{Xincao Xu}
(Member, IEEE) received the B.E. degree from the North University of China, Taiyuan, China, in 2017, and the Ph.D. degree from the Chongqing University, Chongqing, China, in 2023. He is currently an Associate Researcher in computer science with the Shenzhen Institute for Advanced Study, University of Electronic Science and Technology of China (UESTC), Shenzhen, China. From 2023 to 2025, he was a Postdoctoral Research Fellow with the Shenzhen Institute for Advanced Study, UESTC. His research interests include edge intelligence, agentic artificial intelligence, and agentic reinforcement learning.
\end{IEEEbiography}

\begin{IEEEbiography}[{\includegraphics[width=1\textwidth]{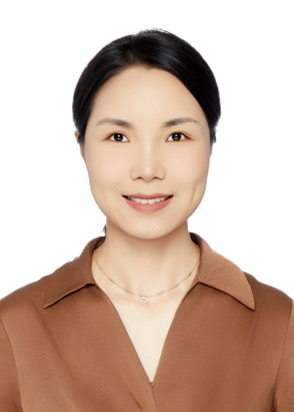}}]{Junhua Wang} (Member, IEEE) received the BS degree and PhD degree in computer science from Chongqing University, in 2014, and 2019, respectively. From 2017 to 2018, she was a visiting scholar in University of Houston, USA. Now she is an associate professor of the school of computer science and engineering, Northeastern University. Her research interests include mobile computing, internet of vehicles, and wireless networks.
\end{IEEEbiography}

\begin{IEEEbiography}[{\includegraphics[width=1\textwidth]{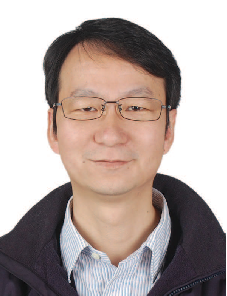}}]
	{Xiao Wu} (S'05-M'08) received the B.Eng. and M.S. degrees in computer science from Yunnan University, Yunnan, China, in 1999 and 2002, respectively, and the Ph.D. degree in Computer Science from City University of Hong Kong, Hong Kong in 2008.
	Currently, he is a Professor and the Assistant Dean of School of Computing and Artificial Intelligence, Southwest Jiaotong University, Chengdu, China. He was with the Institute of Software, Chinese Academy of Sciences, Beijing, China, from 2001 to 2002. He was a Research Assistant and a Senior Research Associate at the City University of Hong Kong, Hong Kong, from 2003 to 2004, and 2007 to 2009, respectively. He was with the School of Computer Science, Carnegie Mellon University, Pittsburgh, PA, USA, and at School of Information and Computer Science, University of California, Irvine, CA, USA as a Visiting Scholar during 2006 to 2007 and 2015 to 2016, respectively. He has authored or co-authored more than 100 research papers in well-respected journals, such as TIP, TMM, TMI and prestigious proceedings like CVPR, ICCV and ACM MM. He received the Second Prize of Natural Science Award of the Ministry of Education, China in 2016 and the Second Prize of Science and Technology Progress Award of Henan Province, China in 2017. His research interests include artificial intelligence, computer vision, multimedia information retrieval, and image/video computing.
\end{IEEEbiography}

\begin{IEEEbiography}[{\includegraphics[width=1\textwidth]{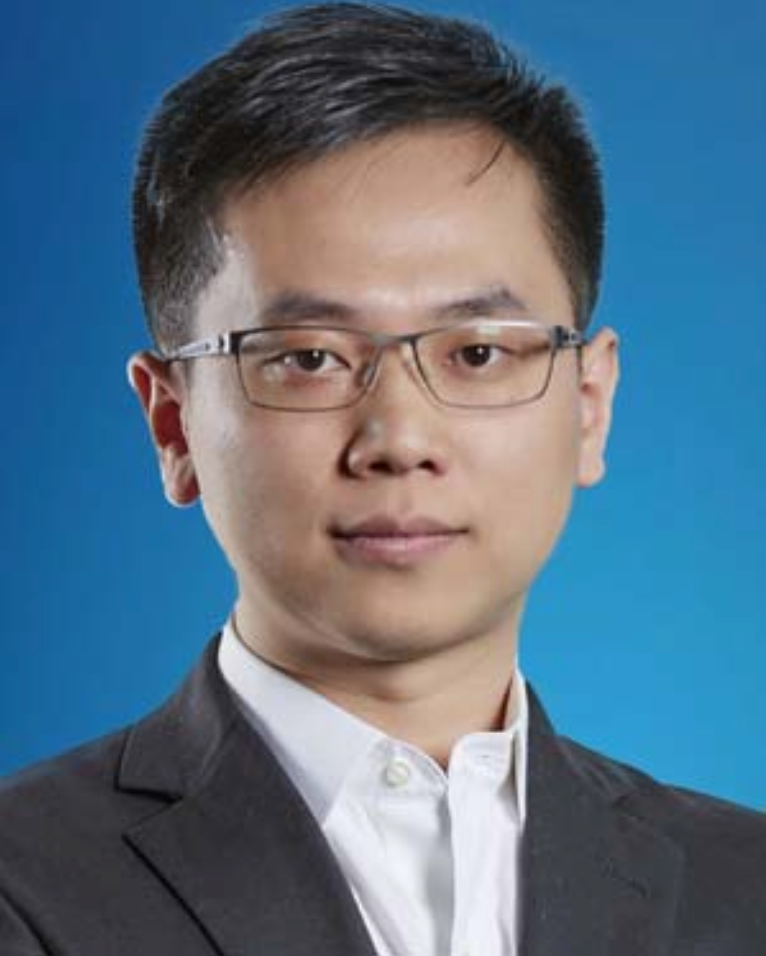}}]{Lixin Duan} ((Member, IEEE) received the B.E. degree from the University of Science and Technology of China, Hefei, China, in 2008, and the Ph.D. degree from Nanyang Technological University, Singapore, in 2012. He is currently a Full Professor with the University of Electronic Science and Technology of China. His current research interests include transfer learning, multiple instance learning, and their applications in computer vision and data mining.
\end{IEEEbiography}

\vfill

\end{document}